\documentclass[nohyperref]{article}
\usepackage[dvipsnames,table,xcdraw]{xcolor}
\usepackage{fullpage}
\usepackage{minitoc}
\usepackage{url}
\usepackage{subfigure}
\usepackage{graphicx}
\usepackage{wrapfig}
\usepackage{amsmath}
\usepackage{amssymb}
\usepackage{microtype}
\usepackage{graphicx}
\usepackage{subcaption}
\usepackage{booktabs}
\usepackage{multirow}
\usepackage{wrapfig}
\usepackage{inputenc}
\usepackage{enumitem}
\usepackage{tcolorbox}
\usepackage{pdflscape}
\usepackage{rotating}
\usepackage{listings}
\usepackage{adjustbox}
\usepackage{comment}
\usepackage{pifont}
\usepackage{booktabs}
\usepackage{multirow}
\usepackage{color, colortbl}
\usepackage[numbers]{natbib}
\usepackage[accsupp]{axessibility}
\usepackage{subfigure}
\usepackage{algpseudocode}
\usepackage{algorithm}

\usepackage[fixed]{fontawesome5}

\usepackage{authblk}
\title{Improving Intervention Efficacy via Concept Realignment \\in Concept Bottleneck Models}

\definecolor{cvprblue}{rgb}{0.21,0.49,0.74}
\usepackage[pagebackref,breaklinks,colorlinks,citecolor=cvprblue]{hyperref}

\usepackage{cleveref}[2012/02/15]%

\crefformat{footnote}{#2\footnotemark[#1]#3}
\Crefname{table}{Table}{Tables}
\crefname{table}{Tab.}{Tabs.}
\Crefname{figure}{Figure}{Figure}
\crefname{figure}{Fig.}{Figs.}
\Crefname{appendix}{Appendix}{Appendix}
\crefname{appendix}{Appx.}{Apps.}
\Crefname{algorithm}{Algorithm}{Algorithm}
\crefname{algorithm}{Alg.}{Algs.}
\Crefname{section}{Section}{Section}
\crefname{section}{Sec.}{Secs.}

\date{}
\makeatletter
\renewcommand\AB@affilsepx{, \protect\Affilfont}
\makeatother
\author{Nishad Singhi$^{1}$\quad Jae Myung Kim$^{1}$\quad Karsten Roth$^{1}$ \quad Zeynep Akata$^{2}$ \\
{\small $^1$T\"ubingen AI Center, University of T\"ubingen \quad $^2$Helmholtz München, TU München} \\
{\small Primary contact: \href{mailto:nishadsinghi@gmail.com}{\texttt{nishadsinghi@gmail.com}}}}

\begin{document}
\maketitle

\doparttoc %
\faketableofcontents %

\vspace{-1.2cm}
\begin{center}
    \begin{tabular}{c@{\hskip 19pt}c}

    % \hspace*{1.4cm}\raisebox{-1pt}{\faGithub} \href{https://github.com/bethgelab/frequency_determines_performance}{\fontsize{8.8pt}{0pt}\path{github.com/bethgelab/frequency_determines_performance}} & \\
    % \hspace*{1.6cm}\raisebox{-1.5pt}{\faDatabase}\href{https://huggingface.co/datasets/bethgelab/Let-It-Wag}{\fontsize{8.8pt}{0pt} \path{huggingface.co/datasets/bethgelab/let-it-wag}} \\
\end{tabular}
\end{center}

\begin{abstract}
%% High-level motivation
\noindent Concept Bottleneck Models (CBMs) ground image classification on human-understandable concepts to allow for interpretable model decisions.
Crucially, the CBM design inherently allows for human interventions, in which expert users are given the ability to modify potentially misaligned concept choices to influence the decision behavior of the model in an interpretable fashion. 
%% Problem Description.
However, existing approaches often require numerous human interventions per image to achieve strong performances, posing practical challenges in scenarios where obtaining human feedback is expensive. 
In this paper, we find that this is noticeably driven by an independent treatment of concepts during intervention, wherein a change of one concept does not influence the use of other ones in the model's final decision. 
%% Proposed Solution.
To address this issue, we introduce a trainable concept intervention realignment module, which leverages concept relations to realign concept assignments post-intervention.
%% Results.
Across standard, real-world benchmarks, we find that concept realignment can significantly improve intervention efficacy; significantly reducing the number of interventions needed to reach a target classification performance or concept prediction accuracy.
In addition, it easily integrates into existing concept-based architectures without requiring changes to the models themselves. 
This reduced cost of human-model collaboration is crucial to enhance the feasibility of CBMs in resource-constrained environments. Our code is available at \href{https://github.com/ExplainableML/concept_realignment}{\texttt{https://github.com/ExplainableML/concept\_realignment}}.
\end{abstract}

\section{Introduction}

Despite tremendous progress of Deep Learning (DL) techniques in research and applications, their adoption to high-stakes scenarios has been limited~\cite{koh2020concept,zarlenga2022concept,zarlenga2023learning}.
This is in large part due to unpredictable biases and failure cases of deep models when transferring to unseen data or complex \& ambiguous cases grounded in the numerous model parameters, architecture designs and training choices ~\cite{dwork2012fairness,chouldechova2016fair,Geirhos2020ShortcutLI,eulig2021diagnose,mehrabi2022survey,dullerud2022is,Brown_2023,roth2024fantastic,casper2024blackbox}. 
The black-box nature of typical DL models and their representation spaces~\cite{shwartzziv2017opening,locatello2019fairness,buhrmester2019analysis,roth2023disentanglement,casper2024blackbox} further exacerbate this problem, as it makes understanding and debugging the decision-making process of these models difficult.
Consequently, it becomes hard for human practitioners to trustworthily operate these models in scenarios with significant legal~\cite{legal1,legal2} or ethical~\cite{ethical1,ethical2} constraints.

\begin{figure}[t]
    \centering
    \includegraphics[width=1\linewidth]{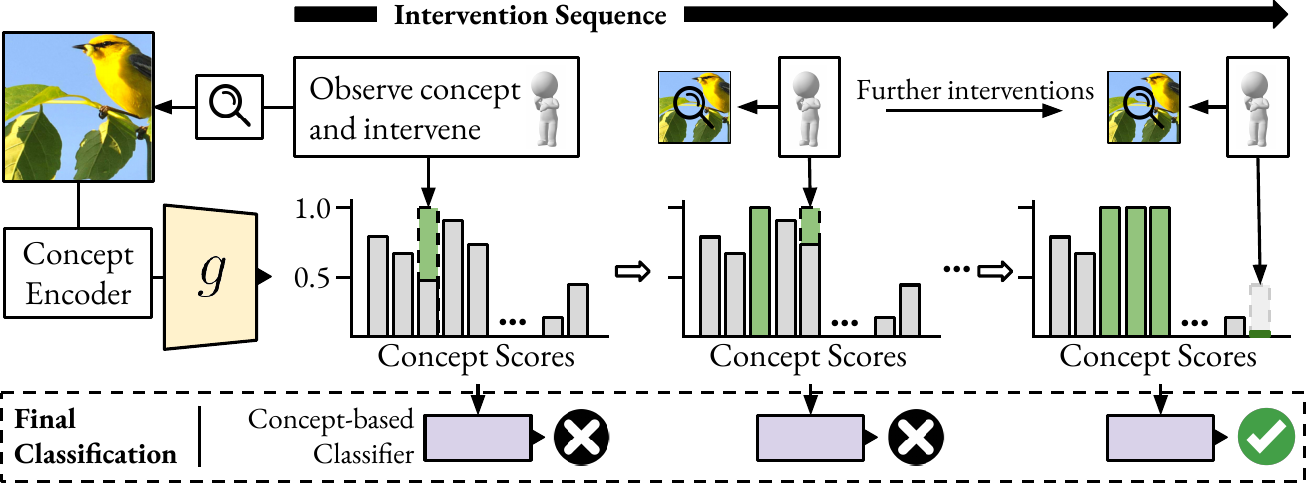}
    \caption{Concept-based classification models allow for \textcolor{ForestGreen}{human intervention}, where a human expert can correct specifically assigned \textcolor{BurntOrange}{concepts}. However, to achieve satisfactory performance, \textcolor{Orchid}{concept-based classification} models often require a large number of interventions, where each additional intervention requires costly human interaction.}
    \label{fig:firstpage}
    \vspace{-10pt}
\end{figure}

To foster trust, transparency in the decision-making process, and the ability to operate alongside expert feedback are required. 
In order to incorporate these desiderata into the design space of deep models, \textit{Koh et al.}~\cite{koh2020concept} introduced Concept Bottleneck Models (CBMs). These models break the decision process into the extraction of human-interpretable concepts (such as "white wings" and "orange beak" when classifying a seagull) from a given input, and a subsequent concept-grounded classifier operating on top of these concept predictions. While this allows users to peek into the model decision process - maybe even more importantly, it also uniquely allows for human-guided intervention and feedback integration at test time. This is done through concept interventions, wherein an expert user analyses predicted concepts and optionally replaces those they deem incorrect with ground-truth information (Fig.~\ref{fig:firstpage}).

Such interventions can significantly raise the performance and reliability of these models~\cite{koh2020concept,zarlenga2022concept,closerlook,zarlenga2023learning}, while offering a natural interface for human-AI collaboration. However, human annotation is expensive, especially when resources and access to expert knowledge are limited. Ideally, such concept models should operate well with minimal human input.
This becomes particularly prevalent as CBMs (as well as follow-up extensions such as Concept Embedding Models (CEMs, \cite{zarlenga2022concept})) often require numerous interventions in order to significantly boost model performance~\cite{koh2020concept,zarlenga2022concept,zarlenga2023learning}, as the set of concepts these models operate on can often be rather extensive.
For example, on the widely used CUB benchmark (bird classification, \cite{wah2011caltech}), it takes 13 interventions per image on overage to raise the accuracy of a baseline CBM model from around $68\%$ to $90\%$ (Fig. \ref{fig:accuracy}).

In this work, we posit that a large part of this limited intervention efficacy can be traced back to the independent nature of how concept interventions are treated. This means that correcting for one concept (or a set of concepts) does not affect which other concepts are predicted for the same image.
However, the occurrence of concepts in real life is often correlated, and informing the model about one concept should consequently influence the use of related ones. 
Not doing so means that we do not leverage human feedback to its full extent - intervening on one specific concept naturally gives additional context about the potential occurrence of other concepts, which should be taken into account in the final classification process.
In particular, we study the extent of this crucial aspect when operating with concept-based models. 
Our study highlights how the use of a \textbf{simple \textit{concept intervention realignment} module}, which learns from statistical concept relations, can effectively and automatically realign concept values after an intervention (or multiple) have been performed. 
Our experiments reveal how our concept intervention realignment can seamlessly integrate into and improve any existing concept-based approach (e.g. default CBMs~\cite{koh2020concept}, advanced CEMs~\cite{zarlenga2022concept} or recently introduced intervention-aware CEMs~\cite{zarlenga2023learning}), and can be deployed both jointly during the initial training of the concept model, and as a post-hoc trained realignment mechanism.
Across three standard, real-world benchmarks (CUB~\cite{wah2011caltech}, CelebA~\cite{liu2015deep} and AwA2~\cite{xian2018zero}), we showcase consistent, in parts very significant improvements in intervention efficacy. 
Across both concept prediction accuracy as well as overall classification accuracy, performance increases more rapidly with interventions as compared to a baseline where concepts are not realigned (in parts reducing the number of interventions needed to reach a target performance by over $70\%$). Combined with its versatile usage and the minimal additional resource requirements, we believe our insights into concept intervention realignment to be of high practical relevance, helping to drive down the cost of human-model collaboration and facilitate the corresponding practical deployment of concept-based models.

\section{Related Works}
% \subsection{Concept Bottleneck Models}
\textbf{Concept Bottleneck Models (CBMs)} have been extensively studied since their introduction by \cite{koh2020concept}. \cite{zarlenga2022concept} proposed Concept Embedding Models as a generalization, utilizing embedding vectors for concepts rather than scalar probabilities, thus enhancing task performance while maintaining interpretability. Recent efforts have explored methods to enhance CBMs without requiring explicit concept supervision during training, leveraging pre-trained vision backbones and language guidance \cite{oikarinen2022label, yang2023language, yuksekgonul2022post}. \cite{alvarez2018towards} introduced Self-explaining Neural Networks (SENNs) for unsupervised concept learning, while \cite{sawada2022concept} proposed CBM-AUC combining SENNs with CBMs. Probabilistic CBMs \cite{probcbm} were proposed to model uncertainty in concepts and final predictions. \cite{mahinpei2021promises, margeloiu2021concept} addressed concept leakage, while techniques proposed by \cite{marconato2022glancenets, havasi2022addressing} aimed to alleviate it. Our work complements these efforts by enabling CBMs to update predictions of all concepts after human intervention.

% \subsection{Interventions on CBMs}
\textbf{Interventions on CBMs.} \cite{koh2020concept} showed that intervening on randomly selected concepts enhances classification performance in CBMs. \cite{chauhan2023interactive} and \cite{sheth2022learning} proposed uncertainty-based strategies for expert interventions. \cite{closerlook} extensively studied concept selection strategies, focusing on task performance and execution cost. \cite{zarlenga2023learning} introduced interventions during training to enhance model receptiveness to test-time interventions. Our approach complements existing methods by updating predictions of all concepts following expert interventions, allowing integration with prior strategies. Concurrently, \cite{xu2023energy} proposed Energy-based CBMs to automate concept prediction updates. In comparison, our method benefits from higher simplicity, improved performance, and seamless integration with existing CBM approaches.
% compared to this concurrent work.
\section{Methods}
\subsection{Background and Preliminaries}\label{subsec:background}

\paragraph{\textbf{Concept Bottleneck Models.}} 
A Concept-Bottleneck Model (CBM) can be viewed as a composition of two models, $h = f(g(x)): \mathcal{X} \rightarrow \mathcal{Y}$, with concept encoder $g: \mathcal{X} \rightarrow \mathcal{C}$, and concept-based classification head $f: \mathcal{C} \rightarrow \mathcal{Y}$. $\mathcal{X}, \mathcal{Y}$, where $\mathcal{C}$ denote input, class label, and concept sets, respectively. 
CBMs get their name from an inherently bifurcated optimization process: While the concept encoder $g(x)$ is trained to predict concepts $\hat{c}\in\mathbb{R}^k$ from the concept set with $\Vert\mathcal{C}\Vert = k$ concepts given an image $x\in\mathbb{R}^d$, the classification head $f(\cdot)$ is optimized to predict final target labels $y \in \mathcal{Y}\in\mathbb{R}^M$ solely based on concept assignments produces by $g$.
CBM training data is thus given as $\mathcal{D}:= \{x^{(i)}, c^{(i)}, y^{(i)}\}_{i = 1}^{N}$, where $x^{(i)}, c^{(i)}, y^{(i)}$ are the inputs, ground-truth concepts, and ground-truth labels, respectively. 
Following existing works~\cite{koh2020concept,zarlenga2022concept,zarlenga2023learning,zarlenga2023towards}, the concept encoder $g$ is trained using a (weighted) binary cross-entropy loss ($\mathcal{L}_{\textrm{concept}} (\hat{c}, c)$), while the classification head $f$ utilizes a cross-entropy classification objective ($\mathcal{L}_{\textrm{task}} (\hat{y}, y) = \mathcal{L}_\text{CE}(\hat{y}, y)$). 

Overall, there are three established schemes~\cite{koh2020concept} for training CBMs: (1) \textit{Independent training:} the concept encoder and classification head are trained entirely independently, with ground-truth concepts $c$ provided as inputs to the classification head during training. (2) \textit{Sequential training:} the concept encoder $g$ is trained first, followed by the classification head $f$ trained using the concepts predicted by $g$. (3) \textit{Joint training:} both the concept encoder $g$ and the classification head $f$ are trained together using a combination of $\mathcal{L}_{\textrm{concept}}$ and $\mathcal{L}_{\textrm{task}}$, respectively.
In all cases, this means that the classification head leverages only information on concept (co-)occurences to predict final class labels, making it easy to ground the final classification decision on interpretable concept assignments.

\paragraph{\textbf{Concept Embedding Models.}}
The flow of information in a CBM is bottlenecked by the set of user-defined concepts. This can potentially limit the processing capacity of the model, especially when the concepts do not contain all the information that is needed to perform the downstream task. To overcome this issue, \cite{zarlenga2022concept} proposed Concept Embedding Models (CEMs) as a generalization of CBMs wherein every concept $i$ is represented by a pair of high-dimensional vectors, $\hat{\textbf{c}}_i^+$ and $\hat{\textbf{c}}_i^-$ (as opposed to scalar concepts in CBMs).
These embeddings are generated by passing $x$ through concept-specific networks  $\phi_i^+$ and $\phi_i^-$, and represent the concept being present and absent, respectively. 

The probability $\hat{p}_i$ of the concept $i$ being in $x$ is then simply computed by passing $\hat{\textbf{c}}_i^+$ and $\hat{\textbf{c}}_i^-$ to a scoring function $s$ as $\hat{p}_i = s([\hat{\textbf{c}}_i^+, \hat{\textbf{c}}_i^-])$. Similarly, both embeddings can also be combined as $\hat{\textbf{c}}_i = \hat{p}_i \hat{\textbf{c}}_i^+ + (1-\hat{p}_i) \hat{\textbf{c}}_i^-$ to parameterize a joint embedding for concept $i$. The final concept embedding which represents the full image $x$ and is passed to the classification head is then given as $\hat{\textbf{c}} := [\hat{\textbf{c}}_1, \hat{\textbf{c}}_2, ..., \hat{\textbf{c}}_k]$. Notice the much higher dimensionality of the concept embedding, which concatenates $k$ concept-specific embeddings (as opposed to producing just a single $k$-dimensional concept vector). 

\paragraph{\textbf{Concept Interventions.}} 
Both CBMs and CEMs allow users to intervene on concepts at test time. Concretely, starting from the concept predictions of the model, $\hat{c}$, the user sequentially intervenes on $T\leq k$ concepts. As a human expert has to both investigate concept predictions and compare against input data, interventions are difficult to parallelize, effectively equating concept intervention into a trajectory of $T$ concept intervention steps~\cite{closerlook} (see also Fig.~\ref{fig:firstpage} for intuition).

Let $\mathcal{S}_t$ represent the set of concepts that have been intervened on up to time $t\leq T$. The corresponding concept embedding at time $t$ is then given as $\tilde{c}_t = \{c_{\mathcal{S}_t}, \hat{c}_{\setminus \mathcal{S}_t} \}$, where $c_{\mathcal{S}_t}$ denotes the ground truth values of the intervened concepts, and $\hat{c}_{\setminus \mathcal{S}_t}$ are the model's predictions of non-intervened concepts. 
Intervening on concepts in this way updates the final prediction of the model from $\hat{y}$ to $\tilde{y} = f(\tilde{c})$. In the case of a CEM, intervening on concept $i$ to update its value from $\hat{p}_i$ to $p_i$ changes its embedding from $\hat{\textbf{c}}_i$ to $\tilde{\textbf{c}}_i = p_i \hat{\textbf{c}}_i^+ + (1-p_i) \hat{\textbf{c}}_i^-$. 

After each intervention $t$, we use a concept intervention policy $\pi(\tilde{c}_t)$ to decide which concept to intervene on next. While $\pi$ can simply suggest random concepts for intervention, it is often much better to leverage heuristics that rank concepts in the order of importance (by some measure). A commonly deployed, effective intervention policy is UCP~\cite{lewis1994heterogeneous,closerlook}, which utilizes the uncertainty of concepts. In particular, UCP selects concepts with the highest uncertainty, i.e. concept predictions closest to 0.5. More details about the intervention process can be found in Algorithm \ref{alg:intervention}.

\paragraph{\textbf{Intervention-aware CEMs.}} 
While test-time interventions typically improve performance, this is not always guaranteed. In fact, recent works have shown that concept interventions can in some cases even hurt the model's performance \cite{zarlenga2023towards, closerlook}. \cite{zarlenga2023learning} noted that this stems from the lack of training incentive for the model to perform well under intervention. 
To address this, they proposed Intervention-aware CEMs (IntCEMs), which introduce interventions during the training process to improve the model's receptiveness to interventions at test time, outperforming all existing methods in the intervention setting. 
In particular, they train a CEM to minimize the following objective:

\begin{equation}
\label{eq:intcem}
\mathcal{L}_{\textrm{IntCEM}}(x, c, y, \mathcal{T}) = \mathcal{L}_{\textrm{pred}}(x, c, y, \hat{c}, \tilde{c}_t) + 
\lambda_{\textrm{conc}} \mathcal{L}_{\textrm{conc}}(\hat{c}, c) + \lambda_{\textrm{roll}} \mathcal{L}_{\textrm{roll}}(x, c, y, \mathcal{T})
\end{equation}
\begin{equation*}
\mathcal{L}_{\textrm{pred}}(x, c, y, \tilde{c}_0, \kappa_t) = \frac{\textrm{CE}(f(\hat{c}, y) + \gamma^T \textrm{CE}(f(\tilde{c}_t), y)}{1 + \gamma^T}
\end{equation*}

$\mathcal{L}_{\textrm{pred}}$ is the prediction loss for $y$, $\mathcal{L}_{\textrm{conc}}$ the concept prediction loss, $\mathcal{L}_{\textrm{roll}}$ the rollout loss incentivizing the model to predict the most informative concept for intervention. $\lambda_{\textrm{conc}}$ and $\lambda_{\textrm{roll}}$ are user-defined weights corresponding to $\mathcal{L}_{\textrm{conc}}$ and $\mathcal{L}_{\textrm{pred}}$ respectively, while $\mathcal{T}$ denotes the intervention trajectory. 
$\mathcal{L}_{\textrm{pred}}$ penalizes the model for incorrect predictions both before and after the intervention, and $\gamma \geq 1$ is a scaling term that prioritizes correct predictions after intervention.

%%%%%%%%%%%%%%%%%%%%%%%%%%%%
\begin{figure}[t]
  \centering
  \includegraphics[width=1\linewidth]{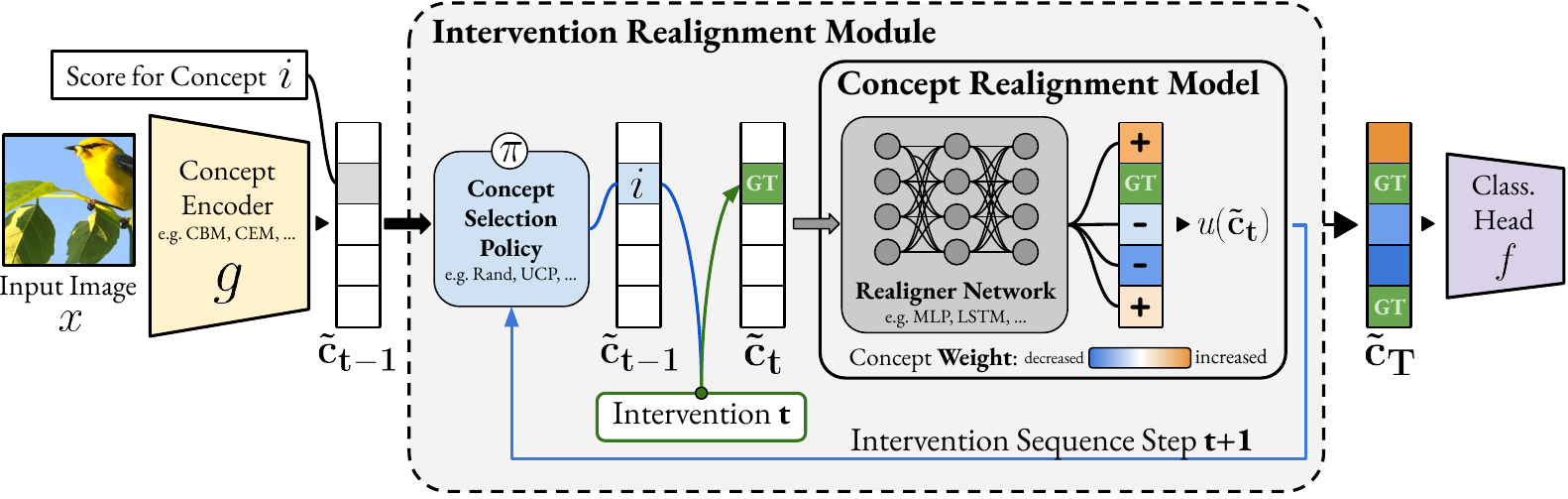} 
  \caption{\textbf{Illustration of the \textcolor{Gray}{concept intervention realignment module}.} Given the \textcolor{BurntOrange}{concept encoding $g(x)$}, we \textcolor{ForestGreen}{intervene} on the concept $i$ selected by a \textcolor{CornflowerBlue}{concept selection policy $\pi$}. This concept is replaced with a ground-truth (GT) value ($\in \{0,1\}$ depending on whether it is present in a given image or not) to obtain $\tilde{c}_t$ (representing intervention step $t \in \{1, ..., T\}$). This intervened concept representation is then passed into the \textcolor{Gray}{concept realignment module} (leveraging e.g. an MLP or LSTM reweighting mode), which outputs the realigned $u(\tilde{c}_t)$. To ensure that the ground-truth values provided by the user are not overwritten during realignment, $u(\tilde{c}_t)$ retains ground-truth corrections. The final concept vector is then based into a \textcolor{Orchid}{concept-based classifier $f$}.}
  \label{fig:diagram}
  \vspace{-10pt}
\end{figure}

%%%%%%%%%%%%%%%%%%%%%%%%%%%%
\subsection{Concept Intervention Realignment}\label{subsec:method}
Previous works incrementally improve on predecessor methods by better parameterizing concept representations or introducing an intervention-aware training objective.
However, all these works still treat concept interventions independently. This means that an intervention on one specific concept has no effect on the assignment of other concepts. 
This disregards relationships between concepts, which in practice do not occur independently (e.g., "white wing" and "white belly" are more likely to co-occur). As a result, the existing intervention process does not utilize human feedback optimally, as information about the verified existence of one concept should naturally guide the prediction of other concepts. While this aspect is naturally important to ensure that an accurate concept representation is passed to the label classifier, it is also crucial when utilizing concept-selection criteria such as UCP because intervening on one concept should consequently reduce the chances of intervening on other closely related, likely co-occurring concepts, while also raising the probability that uncertain and unrelated concepts get intervened on.

\paragraph{\textbf{Intervention Realignment Module.}}
To address this, we propose a \textcolor{Gray}{\textbf{\textit{concept intervention realignment module}}} (\textbf{CIRM}), which consists of two interdependent components: (a) a \textcolor{Gray}{\textbf{\textit{concept realignment model}}} (\textbf{CRM}), $u: \mathcal{C} \rightarrow \mathcal{C}$. After a user intervenes on a subset of concepts $\mathcal{S}$, the remaining concepts ($\setminus\mathcal{S}$) are updated by a realigner network; and (b) an \textcolor{CornflowerBlue}{\textbf{\textit{intervention policy}}} $\pi$. The concepts predicted by the realignment model are fed to the policy to suggest which concept to intervene on next. Both components are interdependent, and together form the overall concept intervention realignment module, as also visualized in Figure~\ref{fig:diagram}.
%
% the use of a simple concept realignment model $u: \mathcal{C} \rightarrow \mathcal{C}$. After a user intervenes on a subset of concepts $\mathcal{S}$, the remaining concepts ($\setminus\mathcal{S}$) are simply updated by the realignment model. Note that the ordering of concepts for intervention typically depends on concept values. Hence, the realignment also directly affects intervention proposals, and thus operates in conjunction with the intervention policy to produce the overall \textcolor{Gray}{\textbf{\textit{intervention realignment module}}}, as visualized in Fig.~\ref{fig:diagram}.
%
The training of the full CIRM comprising both selection policy and concept realignment model aims to simulate the complete intervention process.
It thus starts from the \textcolor{BurntOrange}{concept predictions of the base model}, $\hat{c}$, where we sequentially \textcolor{ForestGreen}{intervene} on concepts for $T \leq k$ time steps by following a policy of choice, $\pi$ (in our case UCP by default, which we experimentally find to outperform random intervention significantly; See Supp. \S \ref{sec:ucp_vs_random}). 

As in \S\ref{subsec:background}, let $\mathcal{S}_t$ denote the set of intervened concepts and $\tilde{c}_t = \{c_{\mathcal{S}_t}, \hat{c}_{\setminus \mathcal{S}_t} \}$ denote the concepts at time $t$, respectively. 
At every intervention time step, we feed $\tilde{c}_t$ to the realignment model to obtain updated concept predictions as $\kappa_t = u(\tilde{c}_t)$, which in turn are utilized by $\pi(\kappa_t)$ to produce intervention recommendations for $t+1$. Finally, we train $u$ with the ground-truth labels as targets using the loss $\mathcal{L}(u) = (\sum_{t = 0}^{T} \textrm{CE} (u(\tilde{c}_t), c))/T$.
%
% \begin{equation}\label{eq:train}
%     \mathcal{L}(u) = \frac{1}{T} \sum_{t = 0}^{T} \textrm{CE} (u(\tilde{c}_t), c)
% \end{equation}
%

Using this simple objective, the concept realignment model $u$ learns to take concept representations and leverage intervened concepts $\mathcal{S}_t$ to predict an updated concept distribution, i.e., $p(i; \hat{c}, \mathcal{S}_t)$. Note that this training objective utilizes standard CBM training information (i.e., concept annotations, \cite{koh2020concept,zarlenga2022concept,zarlenga2023towards,zarlenga2023learning,closerlook}); so no additional information beyond the standard CBM pipeline is required. 

The overall training pipeline can still follow the standard CBM training paradigms (see previous section), with the intervention realignment module being trained independently on top of a pre-trained frozen CBM/CEM as a posthoc realignment method, or jointly with the CBM/CEM to introduce an explicit realignment objective during training. 
For posthoc realignment, we first train the backbone $f$ and the classification head $g$. Subsequently, we freeze those components and train the realignment model $u$.

\paragraph{\textbf{Realignment Models.}} 
As shown in Fig. \ref{fig:diagram}, we parameterize our concept realignment model with a neural network $v$.
To ensure that $u$ does not overwrite the ground-truth concepts provided by the user, we also keep track of the already intervened concepts $\mathcal{S}_t$. Using this information, we replace the output of the realigned concept embedding with the user-provided values for concepts in $\mathcal{S}_t$. Hence, the final output of $u$ for the $i^{th}$ concept is given as 
\[ 
u(\tilde{c}_t, \mathcal{S}_t)^{(i)} = 
\begin{cases}
    v(\tilde{c}_t)^{(i)} & \text{if } i \notin \mathcal{S}_t \\
    \tilde{c}_t^{(i)} & \text{if } i \in \mathcal{S}_t.
\end{cases}
\]
Depending on the assumptions made on the realignment process, $v$ is either a simple MLP or a recurrent model (such as an LSTM~\cite{hochreiter1997long}).
The former parametrizes our default concept intervention realignment model, which only passes the set of intervened and un-intervened concepts at intervention step $t$ to the concept realignment model consisting of a simple MLP. The set of concepts fed into the MLP may either be the original concept embedding $\tilde{c}_0$, where all intervened concepts up to and including step $t$ have been replaced with ground-truth values, or the previously realigned $\kappa_{t-1}$ with similarly updated intervened concepts (c.f. Fig.~\ref{fig:diagram}, "GT"). Note that in either case, $\kappa_{t-1}$ informs the selection process of the subsequent concept to intervene on. After all interventions, the final concept embedding fed into the classifier is always $u(\tilde{c}_T)$. Practically, we found using $\tilde{c}_t$ to work slightly better than $\kappa_{t-1}$. Both cases above however only pass the final set of concepts at time $t$ to the realignment model. Given the sequential nature of interventions, however, it may also be beneficial to account for the entire intervention history to inform future concept realignment. As a result, we also introduce a recurrent realignment variant, $u_\text{rec}$, which employs an LSTM model to retain the entire history of interventions until time $t$. An algorithmic summary is provided in supplementary \S \ref{sec:algorithms}.

\paragraph{\textbf{End-to-End Realignment.}}
In order to jointly train the CIR module and the base model $f \circ g$, we will perform interventions while also training the base model. This naturally combines with the IntCEM framework~\cite{zarlenga2023learning}, which incorporates train-time interventions, and as such is our default choice for joint model and realignment module training. 

Concretely, we modify IntCEMs such that after $t$ interventions, concepts $\tilde{c}_t$ are corrected post-intervention to obtain $\kappa_t = u(\tilde{c}_t)$, which is then fed to the classifier $f$. The new training objective, then, is:
\begin{equation}\label{eq:intcem_loss}
\mathcal{L}_{\textrm{IntCEM-ReA}}(x, c, y, \mathcal{T}) = \mathcal{L}_{\textrm{pred}}(x, c, y, \tilde{c}_0, \kappa_t) + 
\lambda_{\textrm{conc}} \mathcal{L}_{\textrm{conc-ReA}}(\hat{c}, c, \kappa_0, \kappa_t) + \lambda_{\textrm{roll}} \mathcal{L}_{\textrm{roll}}
\end{equation}
\begin{equation}
\mathcal{L}_{\textrm{conc-ReA}}(\hat{c}, c, \kappa_0, \kappa_t) = \frac{1}{2} \left( \mathcal{L}_{\textrm{conc}}(\hat{c}, c) + \frac{\textrm{CE}(\kappa_0, c) + \gamma^T \textrm{CE}(\kappa_T, c)}{1 + \gamma^T} \right) 
\end{equation}
where $\mathcal{L}_{\textrm{conc-ReA}}$ is the modified concept prediction loss which trains both the backbone $g$ of the base model (first term) as well as the CRM (second term). We use the same $\gamma$ as in $\mathcal{L}_{\textrm{pred}}$ to prioritize correct predictions by the CRM after intervention, and the same $\lambda_{\textrm{conc}}$ and $\lambda_{\textrm{roll}}$ as in Eq. \ref{eq:intcem}.

\section{Experiments}
\subsection{Preliminaries}\label{subsec:exp_prelims}

\paragraph{\textbf{Datasets.}} We perform experiments on three datasets:
(1) \textbf{Caltech-UCSD Birds-200-2011 (CUB)} \cite{wah2011caltech} containing $n = 11,788$ bird images over 200 classes. Following the original CBM paper \cite{koh2020concept}, we use 112 concepts grouped into 28 concept groups with the same splits.
(2) \textbf{Large-scale CelebFaces Attributes (CelebA)} \cite{liu2015deep} contains over 200,000 celebrity images annotated with 40 attribute labels, including noisy characteristics such as gender and age. Following \cite{zarlenga2022concept,zarlenga2023learning}, we use only the most balanced 8 concepts in our experiments, resulting in $2^8 = 256$ classes. 
(3) \textbf{Animals with Attributes 2 (AwA2)} \cite{xian2018zero} is a collection of $n = 37,322$ animal images over 50 classes annotated with 85 attributes such as species, color, and behavior.

\paragraph{\textbf{Implementation Details.}}
We perform experiments on CEMs, IntCEMs, and three types of CBMs  (sequential, independent, and joint). For all models and datasets, we follow the hyperparameters used in \cite{zarlenga2023learning}.
During CIRM training, we sequentially intervene on concepts $T = k$ times. By default, we use UCP both during training and inference, and if not stated otherwise, use a multi-layered perceptron (MLP) for concept realignment. 
We use the predictions of the base CBM ($\tilde{c}_t)$ as its input for un-intervened concept representations. We perform a small, standard hyperparameter using Optuna \cite{akiba2019optuna} with 50 trials to search over the number of hidden layers $\in \{1, 2, 3\}$ and units $\in \{k, 2k, k/2\}$, the learning rate $\in [10^{-5}, 10^{-1}]$ and weight decay $\in [10^{-6}, 5 \times 10^{-5}]$, and use the same batch size as used to train the base model. We employ early stopping and learning rate decay on the validation loss. For joint training, we instantiate the realigner MLP 2 hidden layers containing $k$ neurons each. Experiments are conducted using PyTorch ~\cite{pytorch}.

%%%%%%%%%%%%%%%%%%%%%%%%%%%%%%%%%%%%%%%%%%%%%%%%%%%%%%%%%%%%%%%%%%%%%%%%
\subsection{Concept Realignment Improves Intervention Efficacy}\label{subsec:exp_realign}

% \if\useneurips0
% \begin{figure}[t]
%     \centering
%     \begin{subfigure}{0.32\textwidth}
%         \centering
%         \includegraphics[width=\linewidth]{figs/concept_loss_cub_seq_cbm.pdf}
%         \label{fig:sub1}
%     \end{subfigure}
%     \begin{subfigure}{0.32\textwidth}
%         \centering
%         \includegraphics[width=\linewidth]
%         {figs/concept_loss_celeba_seq_cbm.pdf}
%         \label{fig:sub2}
%     \end{subfigure}
%     \begin{subfigure}{0.33\textwidth}
%         \centering
%         \includegraphics[width=\linewidth]
%         {figs/concept_loss_awa2_seq_cbm.pdf}
%         \label{fig:sub3}
%     \end{subfigure}
%     \vspace{-15pt}
%     \caption{Concept prediction loss vs. the number of intervened concepts with and without concept realignment. Concept realignment consistently improves concept predictions.}
%     \label{fig:concept_loss}
% \vspace{-10pt}    
% \end{figure}

% \else
\begin{figure}[t]
    \centering
    \begin{subfigure}
        \centering
        \includegraphics[width=0.31\linewidth]{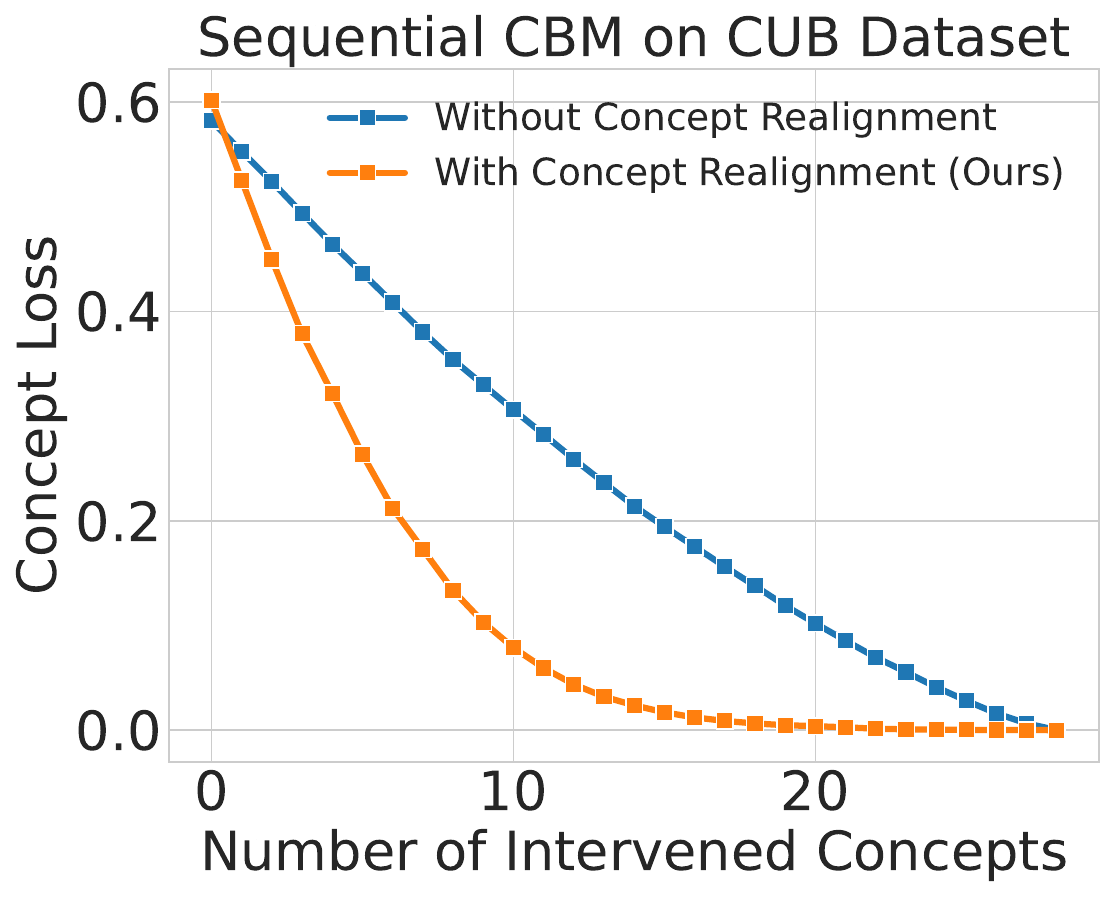}
        \label{fig:sub1}
    \end{subfigure}
    \begin{subfigure}
        \centering
        \includegraphics[width=0.31\linewidth]
        {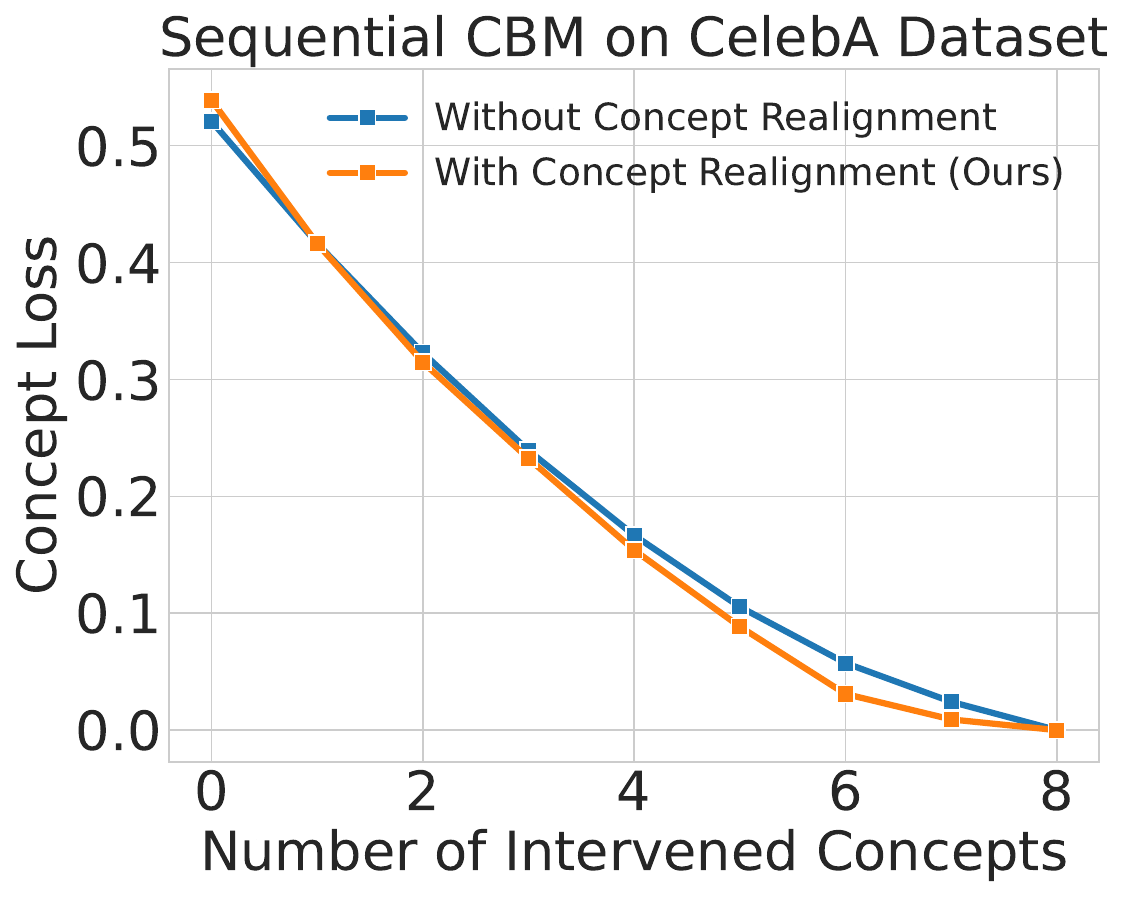}
        \label{fig:sub2}
    \end{subfigure}
    \begin{subfigure}
        \centering
        \includegraphics[width=0.315\linewidth]
        {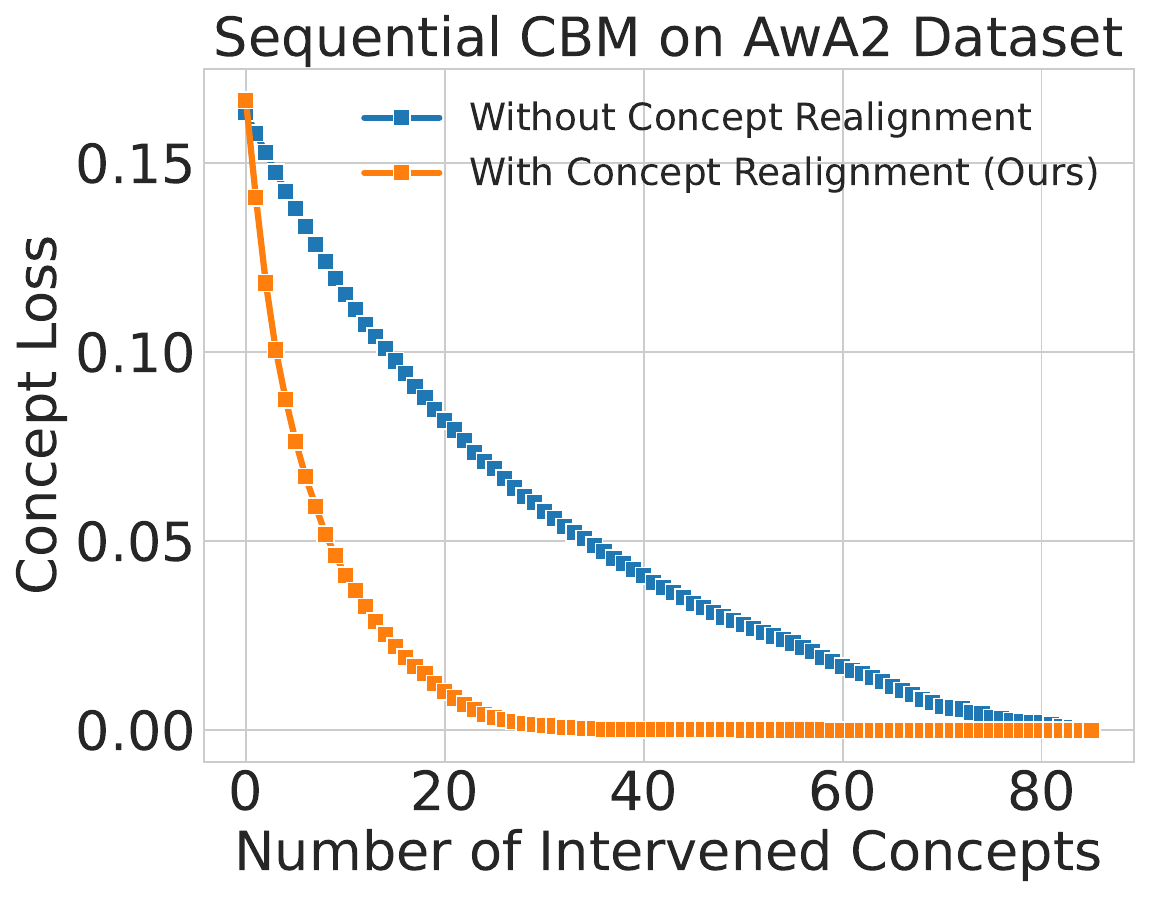}
        \label{fig:sub3}
    \end{subfigure}
    \caption{Concept prediction loss vs. the number of intervened concepts with and without concept realignment. Concept realignment consistently improves concept predictions.}
    \label{fig:concept_loss}
\end{figure}

% \fi

% \if\useneurips0
% \begin{figure}[t]
%     \centering
%     \begin{subfigure}{0.32\textwidth}
%         \centering
%         \includegraphics[width=\linewidth]{figs/accuracy_cub_seq_cbm.pdf}
%         \label{fig:sub1}
%     \end{subfigure}
%     \begin{subfigure}{0.32\textwidth}
%         \centering
%         \includegraphics[width=\linewidth]{figs/accuracy_celeba_seq_cbm.pdf}
%         \label{fig:sub2}
%     \end{subfigure}
%     \begin{subfigure}{0.33\textwidth}
%         \centering
%         \includegraphics[width=\linewidth]{figs/accuracy_awa2_seq_cbm.pdf}
%         \label{fig:sub2}
%     \end{subfigure}
%     \vspace{-15pt}
%     \caption{Classification accuracy vs. the number of intervened concepts with and without concept realignment. Realignment consistently improves classification accuracy.}
%     \vspace{-10pt}    
%     \label{fig:accuracy}
% \end{figure}

% \else
\begin{figure}[t]
    \centering
    \begin{subfigure}
        \centering
        \includegraphics[width=0.31\linewidth]{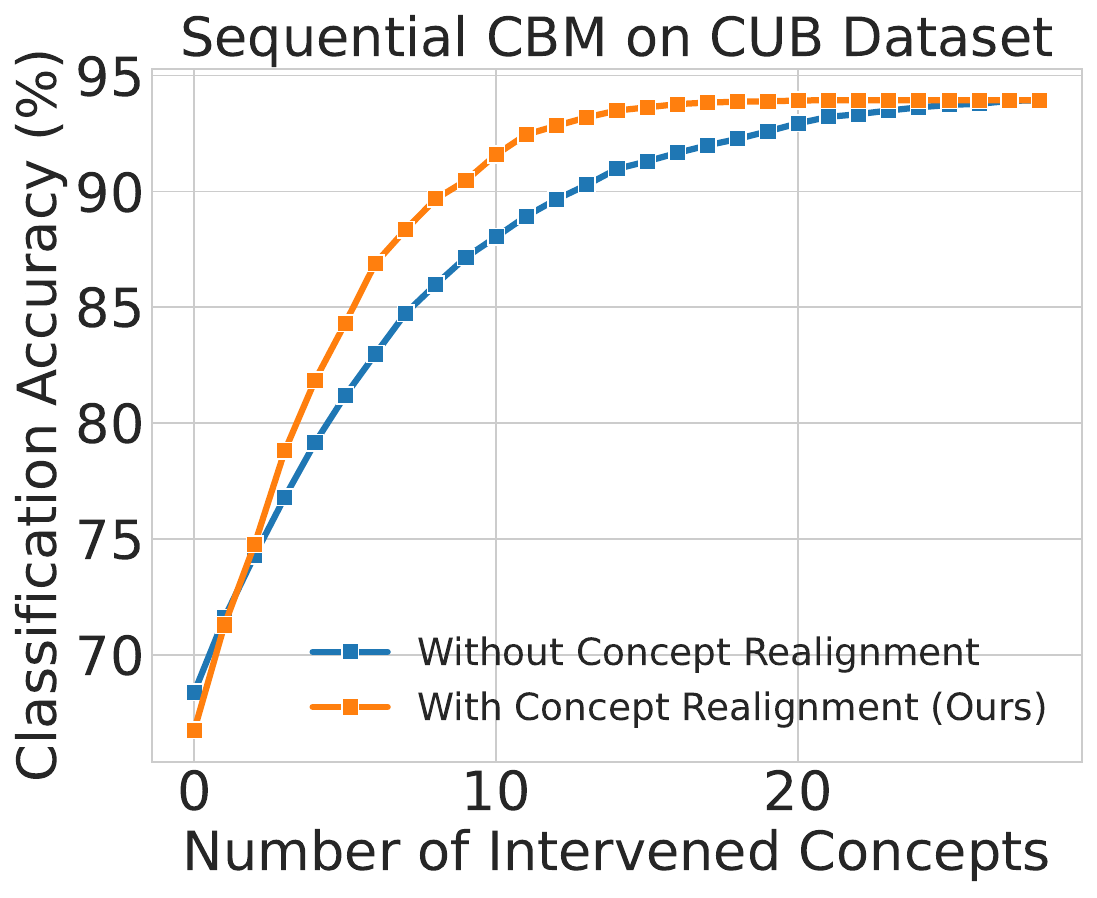}
        \label{fig:sub1}
    \end{subfigure}
    \begin{subfigure}
        \centering
        \includegraphics[width=0.31\linewidth]{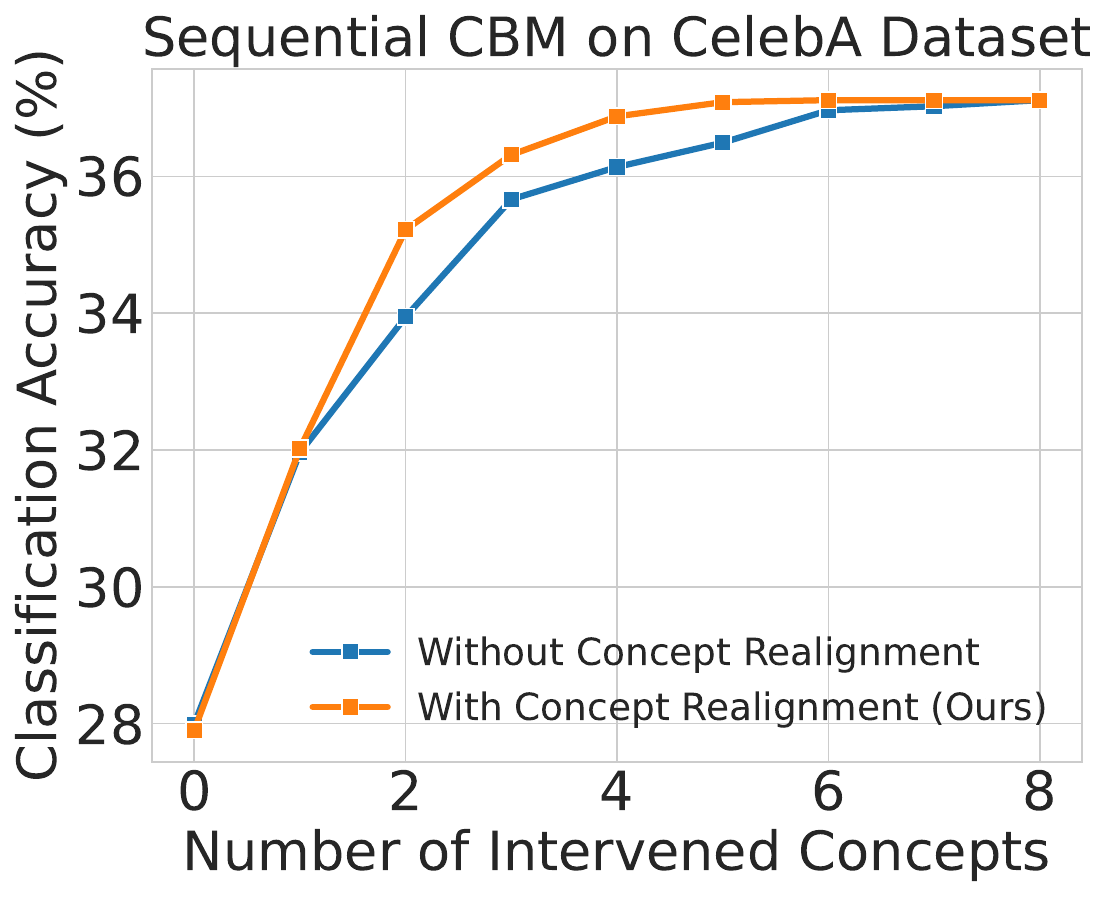}
        \label{fig:sub2}
    \end{subfigure}
    \begin{subfigure}
        \centering
        \includegraphics[width=0.315\linewidth]{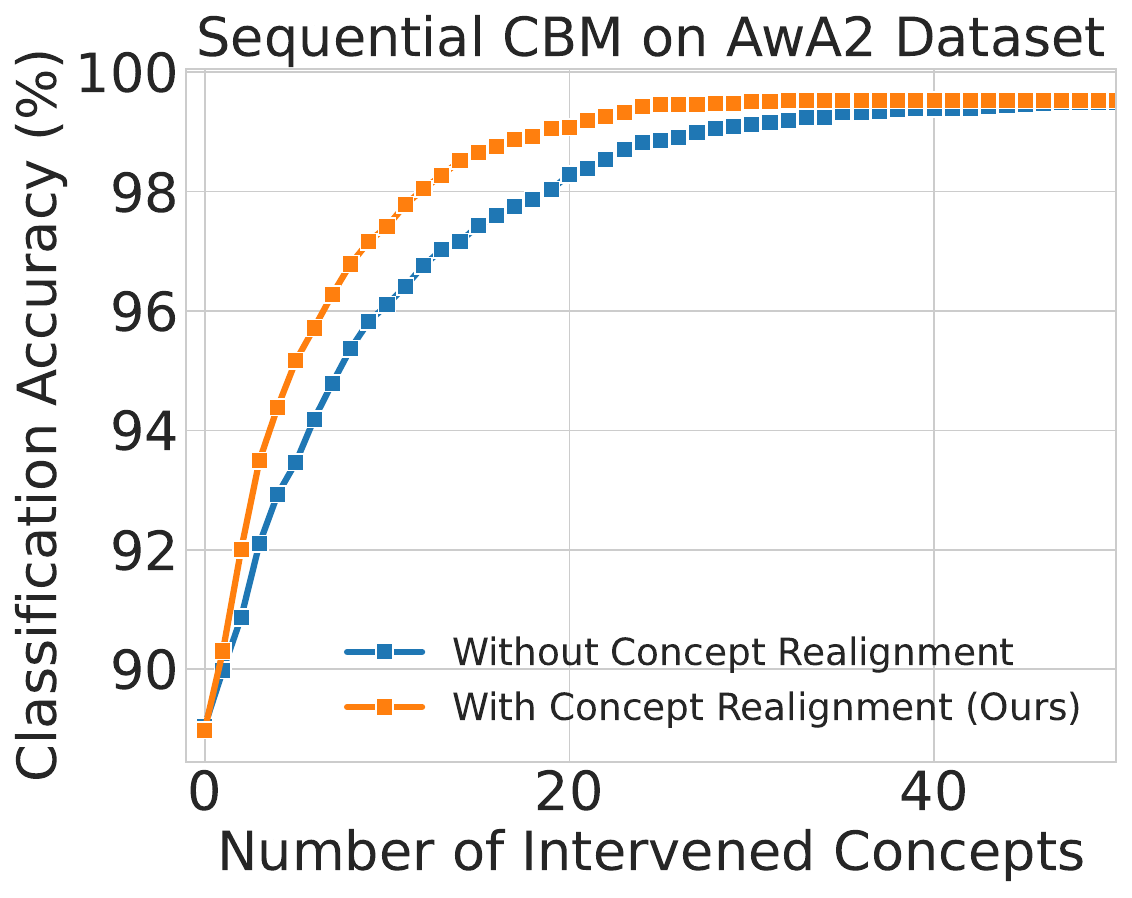}
        \label{fig:sub2}
    \end{subfigure}
    \caption{Classification accuracy vs. the number of intervened concepts with and without concept realignment. Realignment consistently improves classification accuracy.}
    \label{fig:accuracy}
\end{figure}

% \fi

\begin{table}[t]
\centering
\caption{Area Under Curve (AUC) of Concept Prediction Loss and Classification Accuracy with/without CIRM. We use the same backbone for sequential and independent CBMs. CIRM improves performance across all models and datasets. Intervention curves share long saturation plateaus for high intervention counts. Accuracy AUC scores are thus saturated, and best combined with performance graphs in Figs.~\ref{fig:concept_loss},~\ref{fig:accuracy}.}
\resizebox{0.8\linewidth}{!}{%
\begin{tabular}{@{}lccccccc@{}}
\toprule
\multirow{2}{*}{\textbf{Base Model}} & \multirow{2}{*}{\textbf{Realigned}\quad} & \multicolumn{3}{c}{\textbf{Concept Loss AUC $\downarrow$}} & \multicolumn{3}{c}{\textbf{Accuracy AUC $\uparrow$}}\\ 
\cmidrule(l){3-8} 
& & CUB & CelebA & AwA2 & CUB & CelebA & AwA2 \\ 
\hline
\multirow{2}{*}{Sequential CBM} & $\mathbf{\times}$ & 6.71 & 1.59 & 4.26 & 2460.8 & 280.7 & 8364.0 \\
& $\mathbf{\checkmark}$ & \textbf{3.15}  & \textbf{1.52}  & \textbf{1.13} & \textbf{2510.9} & \textbf{284.3} & \textbf{8397.6} \\ \hline
\multirow{2}{*}{Independent CBM} & $\mathbf{\times}$ & 6.71 & 1.59 & 4.26 & 2653.4 & 280.2 & 8403.4 \\
& $\mathbf{\checkmark}$ & \textbf{3.15}  & \textbf{1.52}  & \textbf{1.13} & \textbf{2678.3} & \textbf{282.1} & \textbf{8437.0} \\ \hline
\multirow{2}{*}{Joint CBM}  & $\mathbf{\times}$ & 5.93 & 3.06 & 4.77 & 2580.3 & 273.1 & 8276.4 \\
& $\mathbf{\checkmark}$    & \textbf{3.67}  & \textbf{1.76}  & \textbf{1.48} & \textbf{2609.0} & \textbf{273.9} & \textbf{8327.4} \\ \hline
\multirow{2}{*}{CEM}        & $\mathbf{\times}$  & 5.99 & 1.61 & 4.90 & 2521.4 & 396.3 & 8429.3  \\
                                     & $\mathbf{\checkmark}$    & \textbf{3.20}  & \textbf{1.46}  & \textbf{1.69} & \textbf{2558.4} & \textbf{400.1} & \textbf{8433.9} \\ \hline
\end{tabular}
}
\label{tab:auc}
\end{table}

To probe the efficacy of our concept intervention realignment module, we evaluate both the change in concept prediction loss as well as overall classification accuracy as a function of intervened concept counts. 
These are visualized in Fig.~\ref{fig:concept_loss} and Fig.~\ref{fig:accuracy} for sequentially trained CBMs (see \S\ref{subsec:background}), respectively, for all benchmark test sets - CUB, CelebA and AwA2. Note that for AwA2, we only show the first 50 interventions for visual clarity, as performance beyond that heavily plateaus since sufficient concepts have been intervened on to perfectly solve the test data. 
Table~\ref{tab:auc} numerically summarizes these results via AUC scores and provides additional scores for independently trained CBMs, jointly trained CBMs as well as Concept Embedding Models (CEMs).
Runs in Tab.~\ref{tab:auc} and Figs.~\ref{fig:concept_loss},~\ref{fig:accuracy} all utilize the stronger UCP concept selection policy as opposed to the weaker random selection policy (\S \ref{sec:ucp_vs_random}) to measure intervention efficacy at the highest level, and train the concept realignment module on top of already trained concept models.

\paragraph{\textbf{Improved concept attribution through intervention.}} Across all datasets, we can observe a consistent, in parts vast reduction in concept prediction loss, which measures the correct assignment of concepts for each input (using the concept loss described in \S\ref{subsec:background}). For example on CUB, a \textit{tenfold} reduction of the original unintervened concept loss ($\sim{}0.6$ to $\sim{}0.06$) can be achieved with half the number of interventions ($11$ with concept realignment, $23$ without). This effect becomes even more prevalent on AwA2, where a tenfold reduction ($\sim{}0.17\rightarrow\sim{}0.017$) is achieved after around $16$ interventions with realignment versus more than $60$ without; marking a more than $70\%$ reduction in intervention efforts. This is also reflected in Tab.~\ref{tab:auc}, where concept loss AUC drops by in parts more than half for CUB and from $4.26$ to $1.13$ on AwA2. We find this significant improvement in concept attribution persists across all CBMs and CEMs, as well as random seed initializations (see Supp. Tab.~\ref{tab:3seeds_cub}, ~\ref{tab:3seeds_celeba} and ~\ref{tab:3seeds_awa2})

We do find that for CelebA with a much more restrictive concept bottleneck than e.g. CUB and AwA2, due to significantly fewer (note that in CUB concepts are already grouped, see \S\ref{subsec:exp_prelims}) and noisier concepts, that the overall gain in concept accuracy is smaller. This is also reflected in the notably weaker performance of the base CBM (c.f. Fig.~\ref{fig:accuracy}, middle - less than $38\%$ accuracy when intervening on \textit{all} concepts), which strongly points towards overall insufficient concept information provided in the CelebA training data.
Overall, however, we find very clear evidence that the concept intervention realignment module allows practitioners to leverage human intervention feedback to a much larger extent to attribute the correct concepts to respective inputs. This means that the subsequent classifier will operate on a much more accurate set of concepts, thereby improving the overall interpretability of the final classification decision.

\paragraph{\textbf{Improved overall classification through intervention.}} On top of that, we also find that the significant gain in intervention efficacy on a concept attribution level also translates to subsequent gains in intervention efficacy for the overall classification performance (Fig.~\ref{fig:accuracy}). For example on CUB, the final classification accuracy after intervening on all concepts is $93.9\%$, which is achieved already after $\sim{}16$ intervention steps. A comparable performance without concept intervention realignment requires nearly complete, $\sim{}24$ intervention steps, marking a $50\%$ increase. The same can be seen on CelebA and AwA2 as well, where the upper-bound performance can be achieved with much fewer interventions (particularly without the need to intervene on \textit{all} concepts).
Even intermediate performance targets are achieved much earlier; a classification accuracy target of e.g. $98\%$ on AwA2 requires only $12$ concept interventions with realignment, while the non-aligned baseline needs $19$ interventions on average.
We find these results to be also reflected numerically in Tab~\ref{tab:auc}, where accuracy AUC increases from e.g. $2460.8$ to $2510.9$ on CUB. We do point towards high numerical saturation given the larger performance plateaus at higher intervention counts, and high starting accuracies (e.g. $\sim{}90\%$ on AwA2). Numerical results are thus best considered alongside the intervention trajectories in Figs.~\ref{fig:concept_loss} and \ref{fig:accuracy}.

Together, our experiments provide strong evidence that concept intervention realignment is crucial to best leverage human feedback in concept-based decision systems; allowing to significantly reduce intervention budgets by in parts over $70\%$ to achieve a desired target performance. These gains can also be achieved \textit{after} concept models have been trained, allowing for versatile applicability.

%%%%%%%%%%%%%%%%%%%%%%%%%%%%%%%%%%%%%%%%%%%%%%%%%%%%%%%%%%%%%%%%%%%%%%%%%%%%%%%%%%%%%%%%
\subsection{Intervention Realignment for Intervention-aware CEMs}\label{subsec:joint}

In this section, we investigate training the CIRM during the training process of an already intervention-regularized concept model; namely the recently proposed, state-of-the-art intervention-aware CEM \cite{zarlenga2023learning} (see also \S\ref{subsec:background}). Following the objective described in Eq.~\ref{eq:intcem_loss}, we operate and train the concept intervention module in conjunction with the intervention objective proposed by \cite{zarlenga2023learning}.

% \if\useneurips0
% \begin{figure}[t]
%     \centering
%     \begin{subfigure}{0.35\textwidth}
%         \centering
%         \includegraphics[width=\linewidth]{figs/concept_loss_cub_intcem.pdf}
%         \caption{} % Subcaption for subfigure (a)
%         \label{fig:joint_training_concept_loss}
%     \end{subfigure}
%     \begin{subfigure}{0.355\textwidth} % Adjusted width to fit (b) label properly
%         \centering
%         \includegraphics[width=\linewidth]{figs/accuracy_cub_intcem.pdf}
%         \caption{} % Subcaption for subfigure (b)
%         \label{fig:joint_training_accuracy}
%     \end{subfigure}
%     \vspace{-5pt}
%     \caption{Concept Intervention Realignment in intervention-aware CEMs. (a) Concept prediction loss and (b) classification accuracy with jointly and post-hoc trained CIRMs. In both cases, significant benefits can be seen, especially for correct concept attribution after intervention - both for jointly and posthoc trained realignment modules.} % Main caption for the whole figure
%     \label{fig:joint_training}
%     \vspace{-10pt}
% % \end{wrapfigure}
% \end{figure}

% \else
\begin{figure}[t]
    \centering
    \begin{subfigure}
        \centering
        \includegraphics[width=0.4\linewidth]{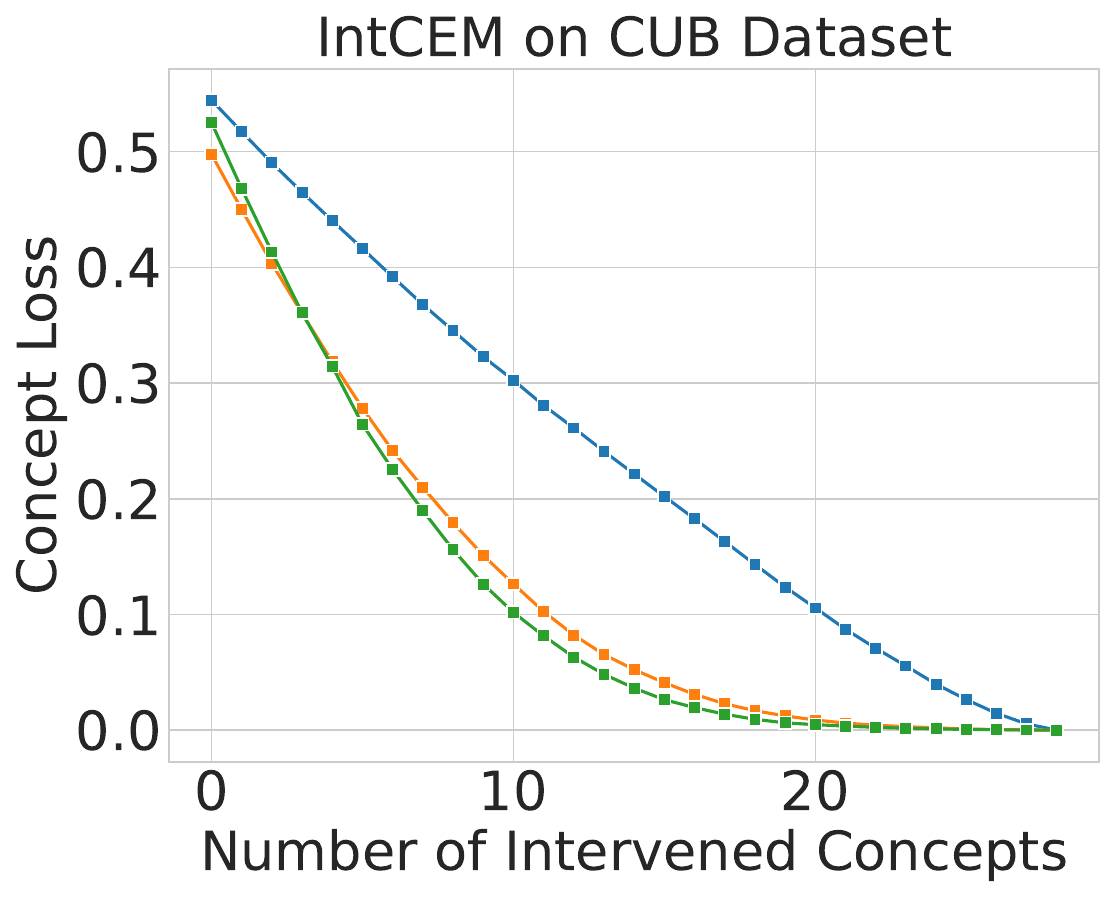}
        \label{fig:joint_training_concept_loss}
    \end{subfigure}
    \begin{subfigure}
        \centering
        \includegraphics[width=0.4\linewidth]{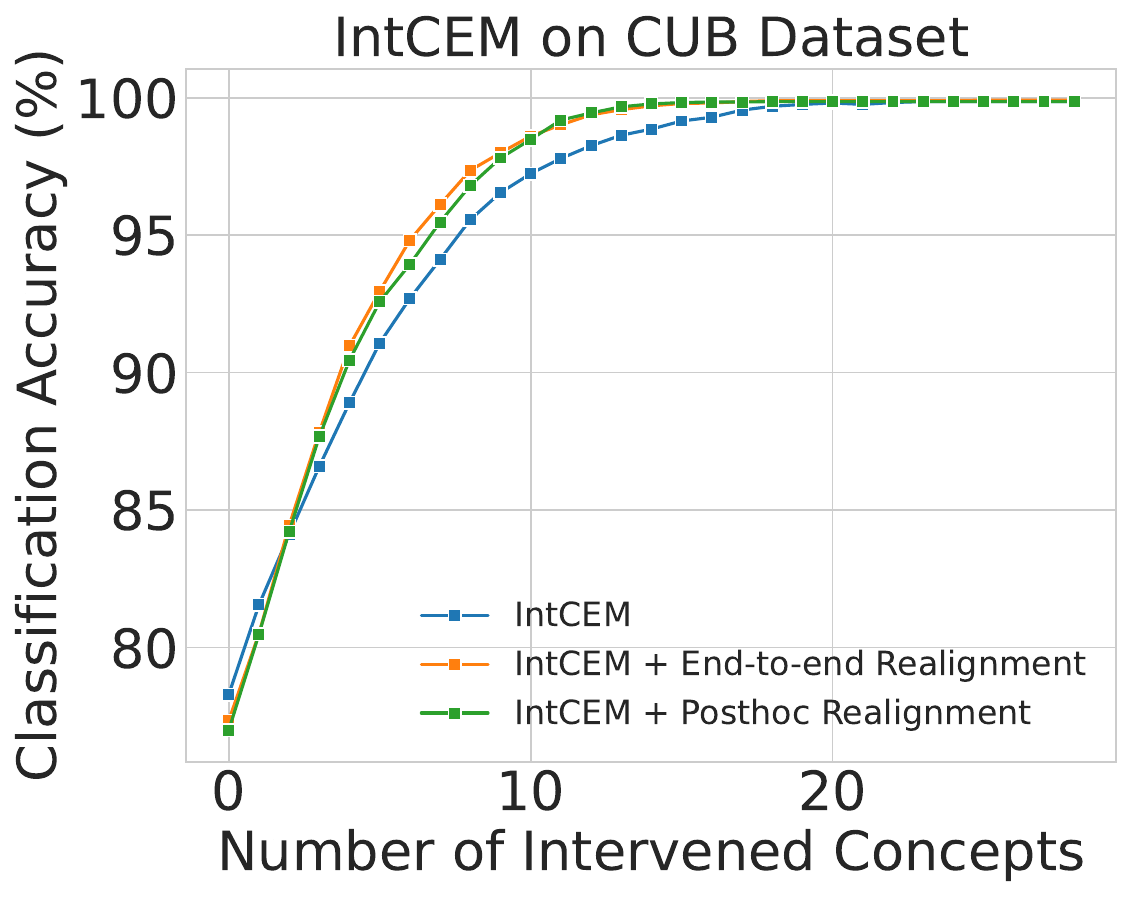}
        \label{fig:joint_training_accuracy}
    \end{subfigure}
    \caption{Concept Intervention Realignment in intervention-aware CEMs. (a) Concept prediction loss and (b) classification accuracy with jointly and post-hoc trained CIRMs. In both cases, significant benefits can be seen, especially for correct concept attribution after intervention - both for jointly and posthoc trained realignment modules.} % Main caption for the whole figure
    \label{fig:joint_training}
% \end{wrapfigure}
\end{figure}

% \fi

Our results are shown in Fig. \ref{fig:joint_training}. First, we find that explicit concept intervention realignment can significantly improve correct concept attribution, even in intervention-aware training setups (c.f. Fig.~\ref{fig:joint_training}a). While not as significant as improvements over standard CBM models, for specific target concept prediction losses (such as a \textit{fivefold} reduction from $0.5$ to $0.1$), half the number of intervention steps are needed ($11$ versus $20$). The improved concept attribution is also reflected in higher intervention accuracies as seen in Fig.~\ref{fig:joint_training}b, albeit the overall (still notable!) improvement is less reflective of the significant gains on a concept level (additional results can be found in Supp. \S \ref{sec:intcem_appendix}). Overall, however, our experiments highlight that even when applied to state-of-the-art approaches that specifically simulate the intervention process during training, improved intervention efficacy can be found. Importantly, the consistently significant improvements on a concept attribution level mean that classification decisions are much better grounded on correct concept attributions, which is crucial for interpretability~\cite{koh2020concept,zarlenga2022concept} of classification results. Finally, we find that concept intervention realignment can be applied both as a regularization mechanism during training, as well as adapted entirely posthoc, while still offering consistent benefits. This supports the high versatility of CIRMs as a general-purpose tool to increase intervention efficacy.

\begin{figure}[t]
    \centering
    \begin{subfigure}
        \centering
        \includegraphics[width=0.4\linewidth]{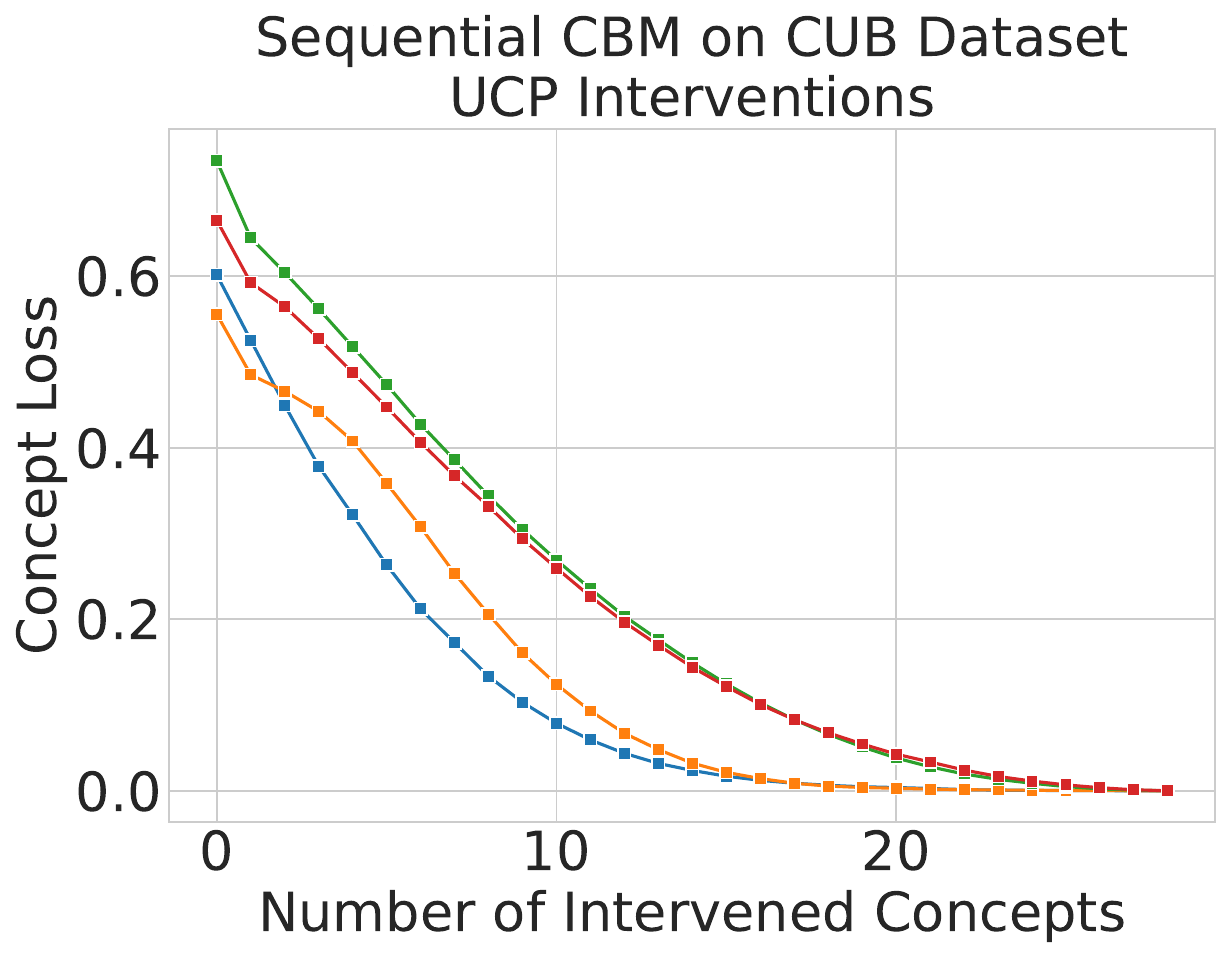}
        \label{fig:ablation_ucp_concept_loss}
    \end{subfigure}
    \begin{subfigure}
        \centering
        \includegraphics[width=0.4\linewidth]{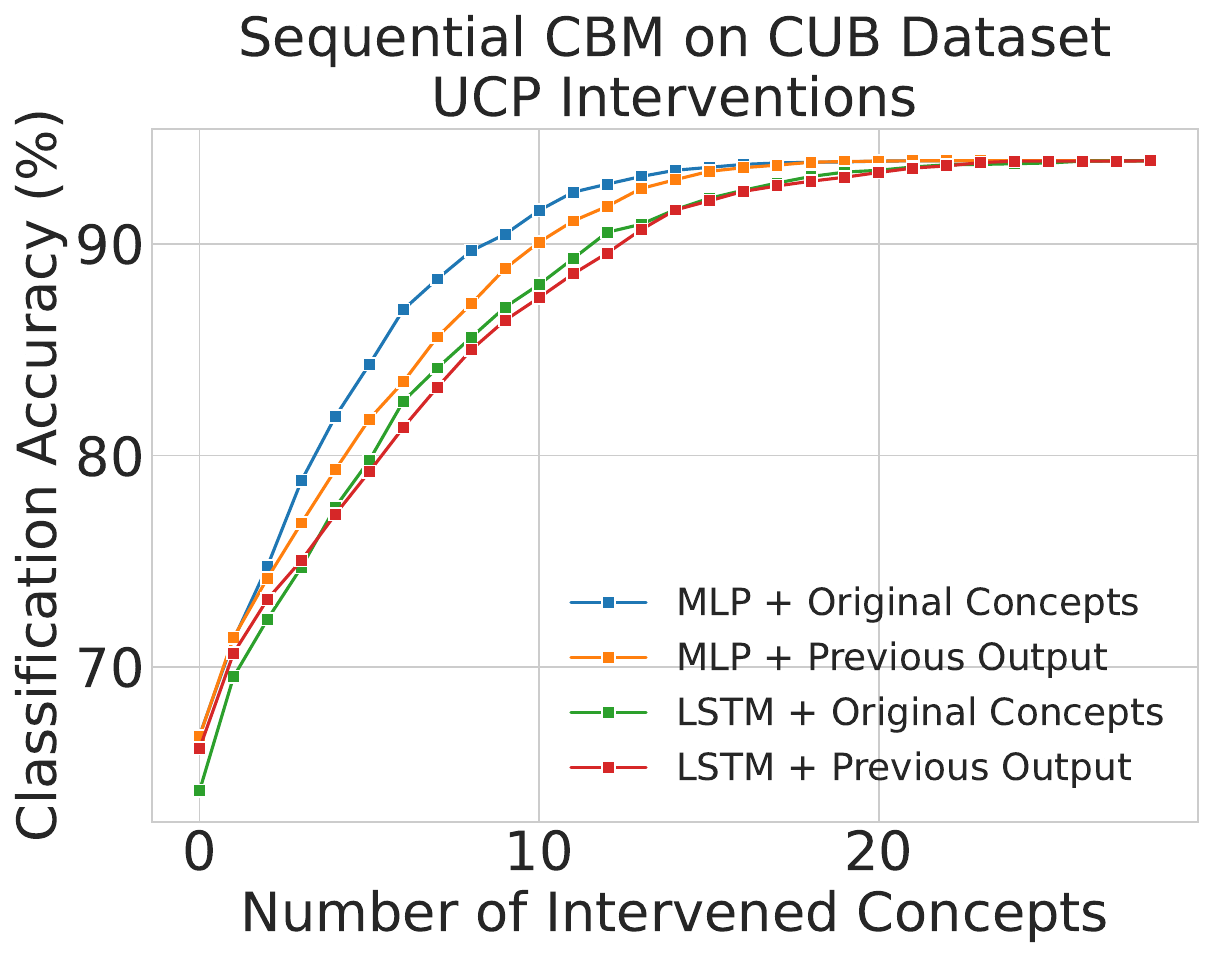}
        \label{fig:ablation_ucp_accuracy}
    \end{subfigure}
    \caption{(a) Concept prediction loss and (b) classification accuracy for various realigner architectures alongside UCP policy. Using an MLP with concept predictions of the base model works better than compounding refinements and accounting for intervention trajectories using LSTMs.}
    \label{fig:ablations}
\end{figure}
% \fi

\subsection{Realignment Module Ablations}
\label{sec:ablations}

\paragraph{\textbf{Realignment Model Architectures.}}

In this section, we study the effect of various design choices for the realignment module along two dimensions:
\textbf{(1) Recurrent vs. Feedforward Networks:} Since we intervene on concepts sequentially, it is possible that the realignment module can benefit from the overall order and history of interventions to make more accurate concept predictions. To do this, we instantiate the concept realignment network using an LSTM \cite{hochreiter1997long}. We compare this against our default MLP.
\textbf{(2) Previous Output vs. Original Concepts:} By default, the realignment module takes as input a combination of ground-truth concepts provided by the user and values predicted by the base model at $t = 0$ for the concepts that have not been intervened on (see also \S\ref{subsec:method}). Due to the sequential nature of interventions, one may also directly feed the output of the realignment module at time $t-1$ as input to it at time $t$ in order to compound the refinements over multiple time steps.
Combining both axes results in four recombinations, which we compare in Fig. \ref{fig:ablations}. As can be seen, there is limited gain when accounting for the complete intervention history using an LSTM realigner network. Similarly, we find that applying the MLP primarily for concept selection alongside UCP and as final input to the classification head works better than compounding refinements over intervention steps.

\paragraph{\textbf{Intervention Policy Transfer.}}

% % \if\useneurips0
% \begin{figure}[t]
%     \centering
%     \begin{subfigure}{0.355\textwidth}
%         \centering
%         \includegraphics[width=\linewidth]{figs/random_interventions_concept_loss.pdf}
%         % \label{fig:sub1}
%     \end{subfigure}
%     \begin{subfigure}{0.35\textwidth}
%         \centering
%         \includegraphics[width=\linewidth]{figs/random_interventions_accuracy.pdf}
%         % \label{fig:sub2}
%     \end{subfigure}
%     % \vspace{-15pt}
%     \vspace{-5pt}
%     \caption{Concept prediction loss and classification accuracy under random interventions for realignment modules trained with random and UCP policy, respectively. Results indicate that alignment of policy used during training and deployment is important.}
%     \label{fig:random_interventions}
%     \vspace{-10pt}
% \end{figure}

% \else
\begin{figure}[t]
    \centering
    \begin{subfigure}
        \centering
        \includegraphics[width=0.4\linewidth]{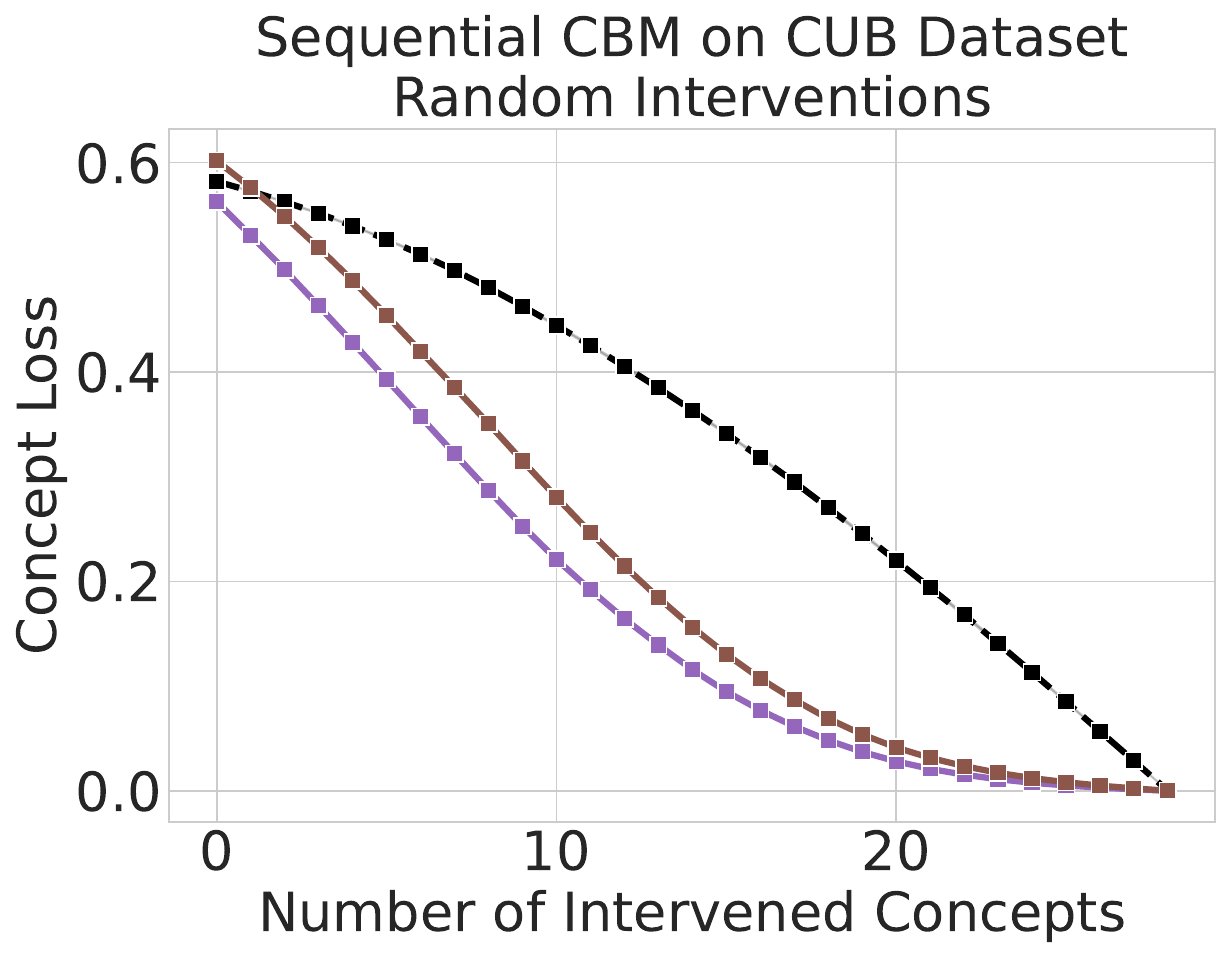}
        % \label{fig:sub1}
    \end{subfigure}
    \begin{subfigure}
        \centering
        \includegraphics[width=0.4\linewidth]{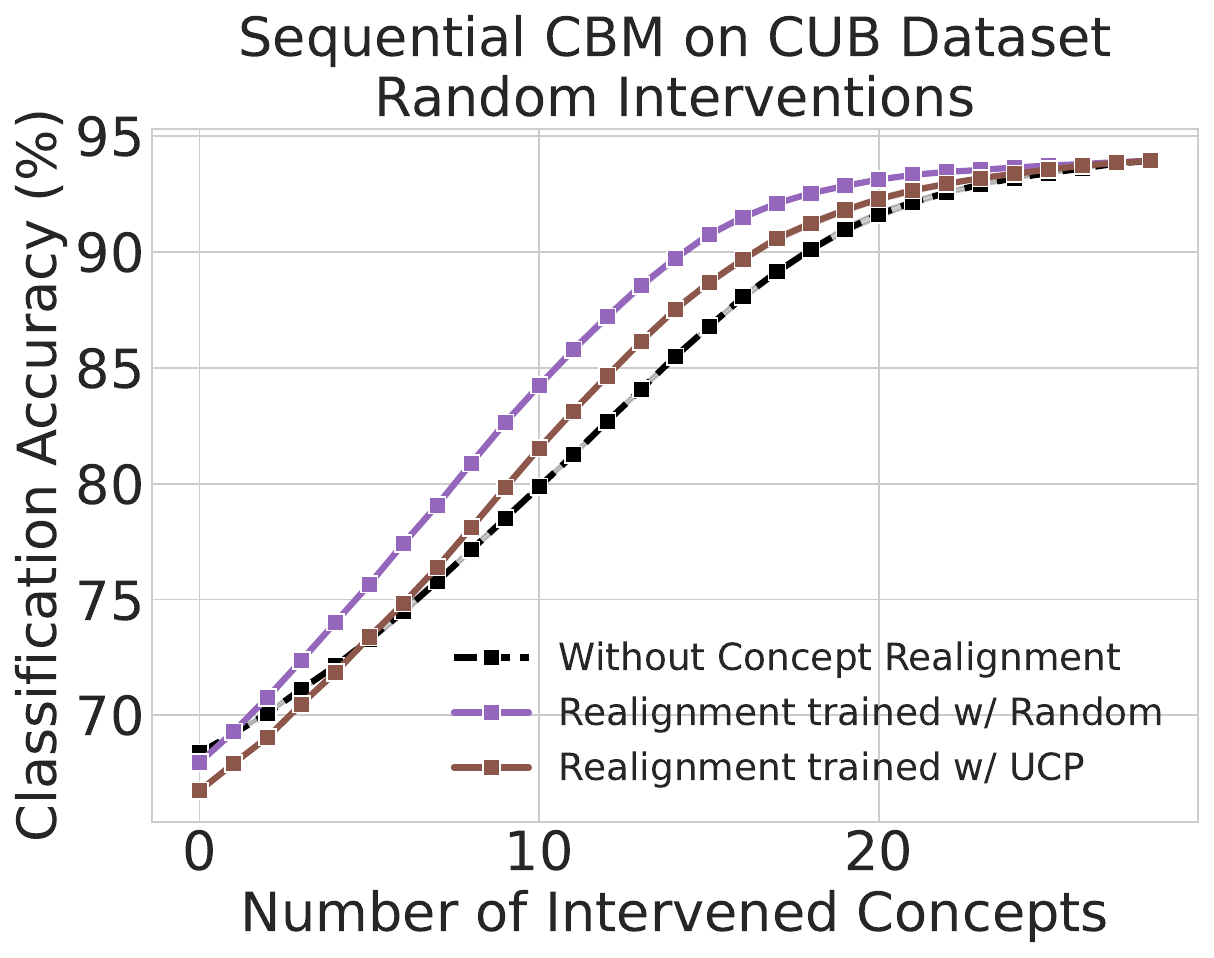}
        % \label{fig:sub2}
    \end{subfigure}
    \caption{Concept prediction loss and classification accuracy under random interventions for realignment modules trained with random and UCP policy, respectively. Results indicate that alignment of policy used during training and deployment is important.}
    \label{fig:random_interventions}
\end{figure}
% \fi

\begin{wrapfigure}{R}{0.4\textwidth}
\vspace{-5pt}
    \centering
        \includegraphics[width=\linewidth]{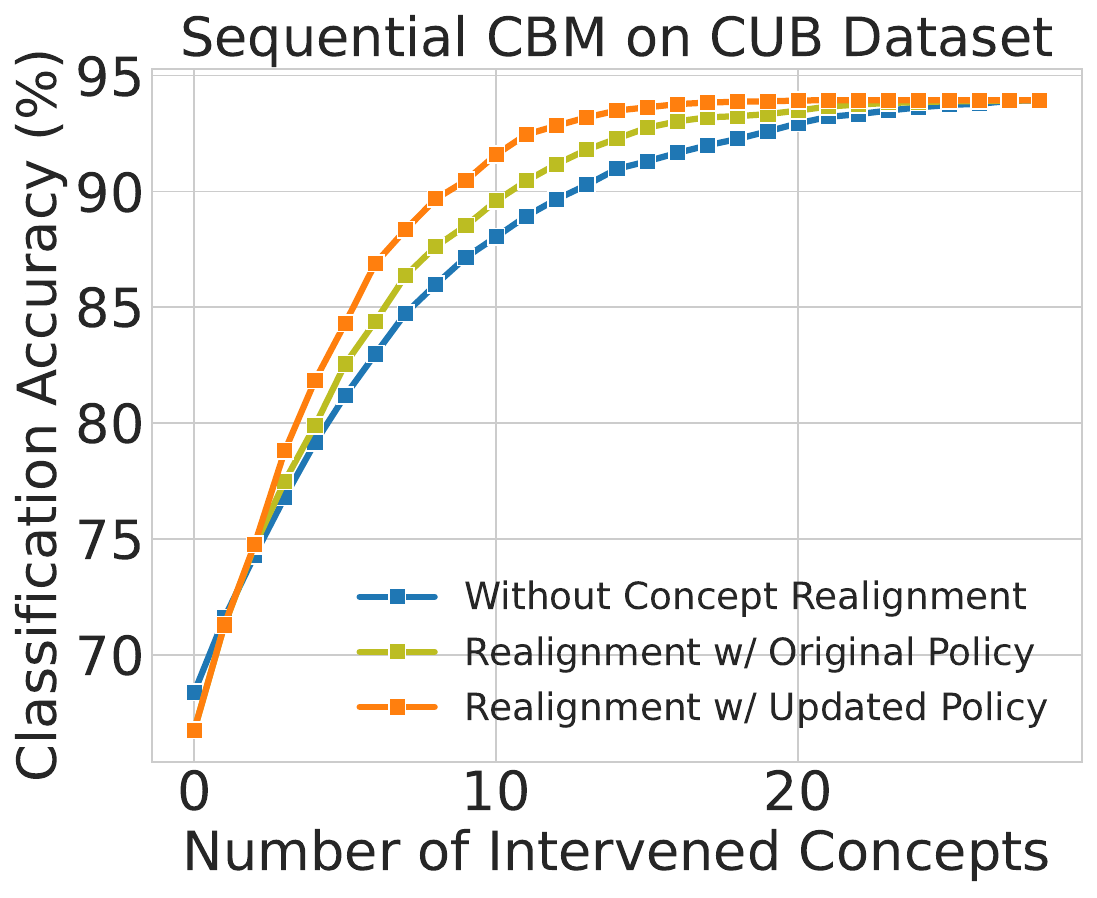}
    \vspace{-15pt}
    \caption{Classification accuracy vs concept interv. counts, showing our updated selection policies improving over the static one.}
    \vspace{-10pt}
    \label{fig:dynamic_vs_static_accuracy}
% \end{figure}
\vspace{-7pt}
\end{wrapfigure}

In this section, we study the importance of aligning intervention policies used during training with those deployed at test time. In particular, we operate on the base setup, which deploys the CBM and the concept intervention realignment module using only the much weaker random intervention policy at test time. However, we change the policy used to train the concept intervention realignment module. Our results are visualized in Fig.~\ref{fig:random_interventions}. As can be seen, while a realignment module trained with UCP can still be effective when deployed with a random intervention policy, it is notably outperformed by the weaker random policy at test-time when the realignment module has been trained on the same random policy as well. This means that the realignment module adapts to the selection policy used during training. Thus to get the most benefits out of concept intervention realignment, selection policies should align during training and deployment.

%%%%%%%%%%%%%%%%%%%%%%%%%%%%%%%%%%%%%%%%%%%%%%%%%%%%%%%%%%%%%%%%%%%%%%%%%%%
\paragraph{\textbf{Alignment b/w Realignment Module Components.}}

Finally, we study how important the alignment between the concept realignment model and intervention policy (i.e., UCP) is to form the overall concept intervention realigment module.
To accomplish this, we employ two module variations: (a) an \textit{original policy} denoted as $\pi(\hat{c}_0)$, which only applies the UCP criterion to the original concept predictions generated by the base model without any concept realignment (i.e., the policy does not change over time), and (b) our default setup (\textit{updated policy}), which informs the intervention policy using realigned concept values ($\pi(\kappa_t)$). Note that in both cases, the classification head still receives realigned concept embeddings, as we only want to study the importance of alignment between the concept realignment model and the intervention policy.
Results in Fig.~\ref{fig:dynamic_vs_static_accuracy} clearly reveal that while simple realignment on its own can already help improve intervention efficacy, much larger efficacy gains are unlocked when both policy and the realignment model are utilized in conjunction.

%%%%%%%%%%%%%%%%%%%%%%%%%%%%%%%%%%%%%%%%%%%%%%%%%%%%%%%%%%%%%%%%%%%%%%%%%%%
\paragraph{\textbf{A Closer Look at Concept Realignment.}}

\begin{figure}[t]
  \centering
  \includegraphics[width=1\linewidth]{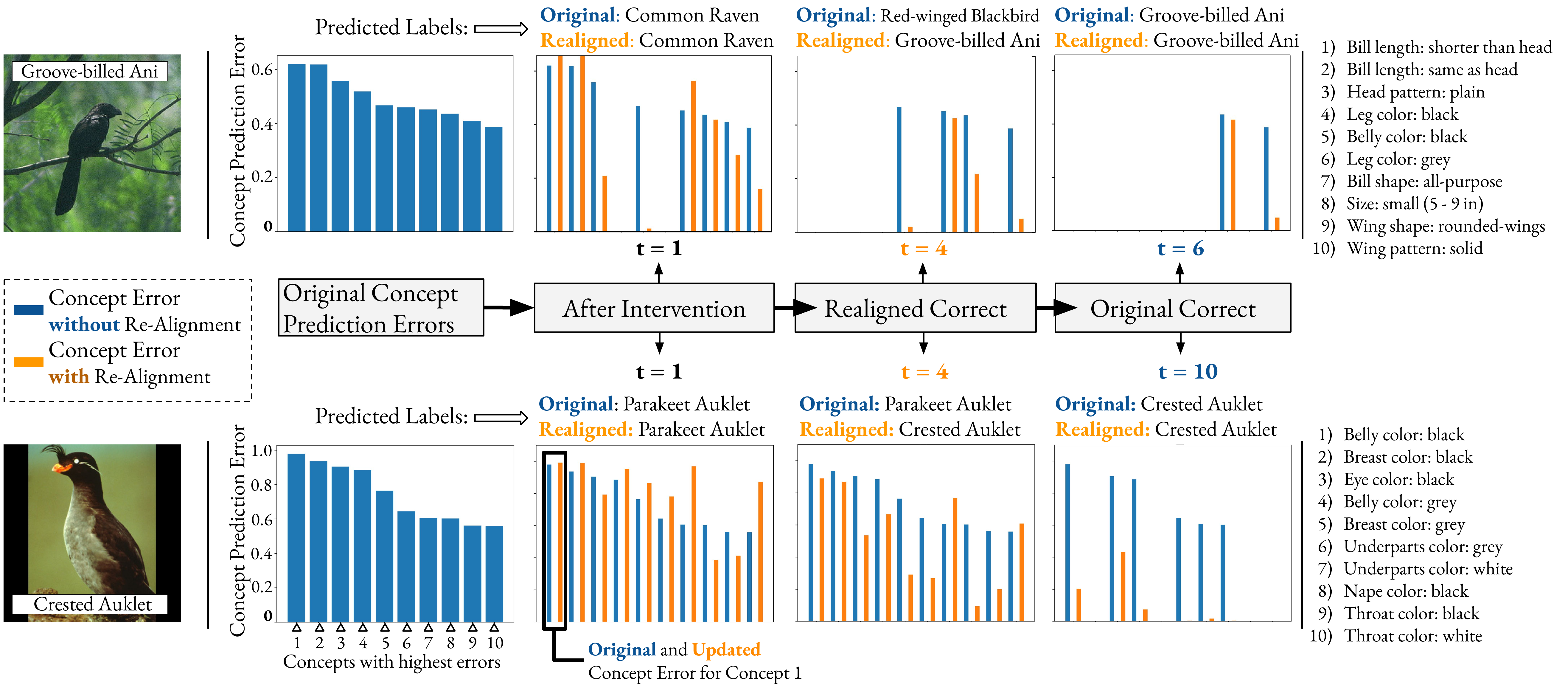} 
  \vspace{-15pt}
  \caption{Qualitative examples for the improved intervention efficiency of CIRM. We show the change in concept prediction errors of the ten worst predicted concepts, as a function of concept intervention steps $t$. As can be seen, concept realignment allows concept error even for strongly mispredicted concepts to be significantly reduced with interventions, achieving correct label classification after much fewer interventions compared to a non-realigned baseline.}
  \label{fig:qualitatives}
  \vspace{-10pt}
\end{figure}

To understand the impact of the realignment process qualitatively, we also provide examples in Fig.~\ref{fig:qualitatives}. In this figure, we showcase the impact of interventions on the top 10 concepts with the highest prediction errors, and the specific number of interventions required to predict the correct label. For both examples, we find that intervention on a single concept is insufficient to flip incorrect class predictions. However, as we intervene on more concepts, we can clearly see that concept realignment jointly allows concept prediction error - even on the initially worst predicted concepts - to be significantly reduced, while also reaching correct image classification with in parts less than half the number of interventions (for \textit{Crested Auklet}). These results conceptually support the quantitative benefits of concept realignment seen in previous benchmark experiments.

\section{Conclusion}
In this work, we identify the independent treatment of concepts during test-time interventions in CBMs as a cause for reduced intervention efficacy. To remedy this problem, we propose a concept intervention realignment module - a simple and lightweight technique to automatically update concept assignments after human intervention on one or multiple concepts. Our experiments demonstrate significant gains in concept attribution as well as overall classification accuracy of concept-based models under intervention. We show that our approach is versatile and can be applied to a wide range of concept-based models, intervention policies, and training schemes. We believe that the reduction in required human interventions to reach performance targets facilitates the practical deployment of concept-based models even in resource-constrained environments.
% Further, we show that updating concepts in this manner after intervention also enables us to select more relevant concepts for intervention, thereby increasing the efficacy of the intervention process.

\section*{Acknowledgements}
Karsten Roth and Jae Myung Kim thank the European Laboratory for Learning and Intelligent Systems (ELLIS) PhD
program and the International Max Planck Research School for Intelligent Systems (IMPRS-IS) for
support. This work was supported by DFG project number 276693517, by BMBF FKZ: 01IS18039A,
by the ERC (853489 - DEXIM), by EXC number 2064/1 – project number 390727645. The authors would like to thank Shyamgopal Karthik and Leander Gerrbach for their helpful feedback on the manuscript.
{
    \small
    \bibliographystyle{plainnat}
    \bibliography{main}

\begin{thebibliography}{42}
\providecommand{\natexlab}[1]{#1}
\providecommand{\url}[1]{\texttt{#1}}
\expandafter\ifx\csname urlstyle\endcsname\relax
  \providecommand{\doi}[1]{doi: #1}\else
  \providecommand{\doi}{doi: \begingroup \urlstyle{rm}\Url}\fi

\bibitem[Akiba et~al.(2019)Akiba, Sano, Yanase, Ohta, and Koyama]{akiba2019optuna}
Takuya Akiba, Shotaro Sano, Toshihiko Yanase, Takeru Ohta, and Masanori Koyama.
\newblock Optuna: A next-generation hyperparameter optimization framework.
\newblock In \emph{Proceedings of the 25th ACM SIGKDD international conference on knowledge discovery \& data mining}, pages 2623--2631, 2019.

\bibitem[Alvarez~Melis and Jaakkola(2018)]{alvarez2018towards}
David Alvarez~Melis and Tommi Jaakkola.
\newblock Towards robust interpretability with self-explaining neural networks.
\newblock \emph{Advances in neural information processing systems}, 31, 2018.

\bibitem[Brown et~al.(2023)Brown, Tomasev, Freyberg, Liu, Karthikesalingam, and Schrouff]{Brown_2023}
Alexander Brown, Nenad Tomasev, Jan Freyberg, Yuan Liu, Alan Karthikesalingam, and Jessica Schrouff.
\newblock Detecting shortcut learning for fair medical ai using shortcut testing.
\newblock \emph{Nature Communications}, 14\penalty0 (1), July 2023.
\newblock ISSN 2041-1723.
\newblock \doi{10.1038/s41467-023-39902-7}.
\newblock URL \url{http://dx.doi.org/10.1038/s41467-023-39902-7}.

\bibitem[Buhrmester et~al.(2019)Buhrmester, Münch, and Arens]{buhrmester2019analysis}
Vanessa Buhrmester, David Münch, and Michael Arens.
\newblock Analysis of explainers of black box deep neural networks for computer vision: A survey, 2019.

\bibitem[Casper et~al.(2024)Casper, Ezell, Siegmann, Kolt, Curtis, Bucknall, Haupt, Wei, Scheurer, Hobbhahn, Sharkey, Krishna, Hagen, Alberti, Chan, Sun, Gerovitch, Bau, Tegmark, Krueger, and Hadfield-Menell]{casper2024blackbox}
Stephen Casper, Carson Ezell, Charlotte Siegmann, Noam Kolt, Taylor~Lynn Curtis, Benjamin Bucknall, Andreas Haupt, Kevin Wei, Jérémy Scheurer, Marius Hobbhahn, Lee Sharkey, Satyapriya Krishna, Marvin~Von Hagen, Silas Alberti, Alan Chan, Qinyi Sun, Michael Gerovitch, David Bau, Max Tegmark, David Krueger, and Dylan Hadfield-Menell.
\newblock Black-box access is insufficient for rigorous ai audits, 2024.

\bibitem[Chauhan et~al.(2023)Chauhan, Tiwari, Freyberg, Shenoy, and Dvijotham]{chauhan2023interactive}
Kushal Chauhan, Rishabh Tiwari, Jan Freyberg, Pradeep Shenoy, and Krishnamurthy Dvijotham.
\newblock Interactive concept bottleneck models.
\newblock In \emph{Proceedings of the AAAI Conference on Artificial Intelligence}, volume~37, pages 5948--5955, 2023.

\bibitem[Chouldechova(2016)]{chouldechova2016fair}
Alexandra Chouldechova.
\newblock Fair prediction with disparate impact: A study of bias in recidivism prediction instruments, 2016.

\bibitem[Dullerud et~al.(2022)Dullerud, Roth, Hamidieh, Papernot, and Ghassemi]{dullerud2022is}
Natalie Dullerud, Karsten Roth, Kimia Hamidieh, Nicolas Papernot, and Marzyeh Ghassemi.
\newblock Is fairness only metric deep? evaluating and addressing subgroup gaps in deep metric learning.
\newblock In \emph{International Conference on Learning Representations}, 2022.
\newblock URL \url{https://openreview.net/forum?id=js62_xuLDDv}.

\bibitem[Dur{\'a}n and Jongsma(2021)]{ethical1}
Juan~Manuel Dur{\'a}n and Karin~Rolanda Jongsma.
\newblock Who is afraid of black box algorithms? on the epistemological and ethical basis of trust in medical ai.
\newblock \emph{Journal of Medical Ethics}, 47\penalty0 (5):\penalty0 329--335, 2021.
\newblock ISSN 0306-6800.
\newblock \doi{10.1136/medethics-2020-106820}.
\newblock URL \url{https://jme.bmj.com/content/47/5/329}.

\bibitem[Dwork et~al.(2012)Dwork, Hardt, Pitassi, Reingold, and Zemel]{dwork2012fairness}
Cynthia Dwork, Moritz Hardt, Toniann Pitassi, Omer Reingold, and Richard Zemel.
\newblock Fairness through awareness.
\newblock In \emph{Proceedings of the 3rd Innovations in Theoretical Computer Science Conference}, ITCS '12, page 214–226, New York, NY, USA, 2012. Association for Computing Machinery.
\newblock ISBN 9781450311151.
\newblock \doi{10.1145/2090236.2090255}.
\newblock URL \url{https://doi.org/10.1145/2090236.2090255}.

\bibitem[EUGDPR(2017)]{legal1}
EUGDPR.
\newblock Gdpr. general data protection regulation, 2017.

\bibitem[Eulig et~al.(2021)Eulig, Saranrittichai, Mummadi, Rambach, Beluch, Shi, and Fischer]{eulig2021diagnose}
Elias Eulig, Piyapat Saranrittichai, Chaithanya~Kumar Mummadi, Kilian Rambach, William Beluch, Xiahan Shi, and Volker Fischer.
\newblock Diagvib-6: A diagnostic benchmark suite for vision models in the presence of shortcut and generalization opportunities.
\newblock In \emph{Proceedings of the IEEE/CVF International Conference on Computer Vision (ICCV)}, October 2021.

\bibitem[Geirhos et~al.(2020)Geirhos, Jacobsen, Michaelis, Zemel, Brendel, Bethge, and Wichmann]{Geirhos2020ShortcutLI}
Robert Geirhos, J{\"o}rn-Henrik Jacobsen, Claudio Michaelis, Richard~S. Zemel, Wieland Brendel, Matthias Bethge, and Felix Wichmann.
\newblock Shortcut learning in deep neural networks.
\newblock \emph{Nature Machine Intelligence}, 2:\penalty0 665 -- 673, 2020.

\bibitem[Havasi et~al.(2022)Havasi, Parbhoo, and Doshi-Velez]{havasi2022addressing}
Marton Havasi, Sonali Parbhoo, and Finale Doshi-Velez.
\newblock Addressing leakage in concept bottleneck models.
\newblock \emph{Advances in Neural Information Processing Systems}, 35:\penalty0 23386--23397, 2022.

\bibitem[Hochreiter and Schmidhuber(1997)]{hochreiter1997long}
Sepp Hochreiter and J{\"u}rgen Schmidhuber.
\newblock Long short-term memory.
\newblock \emph{Neural computation}, 9\penalty0 (8):\penalty0 1735--1780, 1997.

\bibitem[Kim et~al.(2023)Kim, Jung, Park, Kim, and Yoon]{probcbm}
Eunji Kim, Dahuin Jung, Sangha Park, Siwon Kim, and Sungroh Yoon.
\newblock Probabilistic concept bottleneck models.
\newblock In Andreas Krause, Emma Brunskill, Kyunghyun Cho, Barbara Engelhardt, Sivan Sabato, and Jonathan Scarlett, editors, \emph{Proceedings of the 40th International Conference on Machine Learning}, volume 202 of \emph{Proceedings of Machine Learning Research}, pages 16521--16540. PMLR, 23--29 Jul 2023.
\newblock URL \url{https://proceedings.mlr.press/v202/kim23g.html}.

\bibitem[Koh et~al.(2020)Koh, Nguyen, Tang, Mussmann, Pierson, Kim, and Liang]{koh2020concept}
Pang~Wei Koh, Thao Nguyen, Yew~Siang Tang, Stephen Mussmann, Emma Pierson, Been Kim, and Percy Liang.
\newblock Concept bottleneck models.
\newblock In \emph{International conference on machine learning}, pages 5338--5348. PMLR, 2020.

\bibitem[Lewis and Catlett(1994)]{lewis1994heterogeneous}
David~D Lewis and Jason Catlett.
\newblock Heterogeneous uncertainty sampling for supervised learning.
\newblock In \emph{Machine learning proceedings 1994}, pages 148--156. Elsevier, 1994.

\bibitem[Liu et~al.(2015)Liu, Luo, Wang, and Tang]{liu2015deep}
Ziwei Liu, Ping Luo, Xiaogang Wang, and Xiaoou Tang.
\newblock Deep learning face attributes in the wild.
\newblock In \emph{Proceedings of the IEEE international conference on computer vision}, pages 3730--3738, 2015.

\bibitem[Locatello et~al.(2019)Locatello, Abbati, Rainforth, Bauer, Sch\"{o}lkopf, and Bachem]{locatello2019fairness}
Francesco Locatello, Gabriele Abbati, Tom Rainforth, Stefan Bauer, Bernhard Sch\"{o}lkopf, and Olivier Bachem.
\newblock \emph{On the fairness of disentangled representations}.
\newblock Curran Associates Inc., Red Hook, NY, USA, 2019.

\bibitem[Mahinpei et~al.(2021)Mahinpei, Clark, Lage, Doshi-Velez, and Pan]{mahinpei2021promises}
Anita Mahinpei, Justin Clark, Isaac Lage, Finale Doshi-Velez, and Weiwei Pan.
\newblock Promises and pitfalls of black-box concept learning models.
\newblock \emph{arXiv preprint arXiv:2106.13314}, 2021.

\bibitem[Marconato et~al.(2022)Marconato, Passerini, and Teso]{marconato2022glancenets}
Emanuele Marconato, Andrea Passerini, and Stefano Teso.
\newblock Glancenets: Interpretable, leak-proof concept-based models.
\newblock \emph{Advances in Neural Information Processing Systems}, 35:\penalty0 21212--21227, 2022.

\bibitem[Margeloiu et~al.(2021)Margeloiu, Ashman, Bhatt, Chen, Jamnik, and Weller]{margeloiu2021concept}
Andrei Margeloiu, Matthew Ashman, Umang Bhatt, Yanzhi Chen, Mateja Jamnik, and Adrian Weller.
\newblock Do concept bottleneck models learn as intended?
\newblock \emph{arXiv preprint arXiv:2105.04289}, 2021.

\bibitem[Mehrabi et~al.(2022)Mehrabi, Morstatter, Saxena, Lerman, and Galstyan]{mehrabi2022survey}
Ninareh Mehrabi, Fred Morstatter, Nripsuta Saxena, Kristina Lerman, and Aram Galstyan.
\newblock A survey on bias and fairness in machine learning, 2022.

\bibitem[Oikarinen et~al.(2022)Oikarinen, Das, Nguyen, and Weng]{oikarinen2022label}
Tuomas Oikarinen, Subhro Das, Lam~M Nguyen, and Tsui-Wei Weng.
\newblock Label-free concept bottleneck models.
\newblock In \emph{The Eleventh International Conference on Learning Representations}, 2022.

\bibitem[Paszke et~al.(2019)Paszke, Gross, Massa, Lerer, Bradbury, Chanan, Killeen, Lin, Gimelshein, Antiga, Desmaison, Kopf, Yang, DeVito, Raison, Tejani, Chilamkurthy, Steiner, Fang, Bai, and Chintala]{pytorch}
Adam Paszke, Sam Gross, Francisco Massa, Adam Lerer, James Bradbury, Gregory Chanan, Trevor Killeen, Zeming Lin, Natalia Gimelshein, Luca Antiga, Alban Desmaison, Andreas Kopf, Edward Yang, Zachary DeVito, Martin Raison, Alykhan Tejani, Sasank Chilamkurthy, Benoit Steiner, Lu~Fang, Junjie Bai, and Soumith Chintala.
\newblock Pytorch: An imperative style, high-performance deep learning library.
\newblock In \emph{Advances in Neural Information Processing Systems 32}, pages 8024--8035. Curran Associates, Inc., 2019.
\newblock URL \url{http://papers.neurips.cc/paper/9015-pytorch-an-imperative-style-high-performance-deep-learning-library.pdf}.

\bibitem[Piano(2020)]{ethical2}
Samuele~Lo Piano.
\newblock Ethical principles in machine learning and artificial intelligence: cases from the field and possible ways forward.
\newblock \emph{Palgrave Communications}, 7\penalty0 (1):\penalty0 1--7, 2020.
\newblock URL \url{https://EconPapers.repec.org/RePEc:pal:palcom:v:7:y:2020:i:1:d:10.1057_s41599-020-0501-9}.

\bibitem[Roth et~al.(2023)Roth, Ibrahim, Akata, Vincent, and Bouchacourt]{roth2023disentanglement}
Karsten Roth, Mark Ibrahim, Zeynep Akata, Pascal Vincent, and Diane Bouchacourt.
\newblock Disentanglement of correlated factors via hausdorff factorized support.
\newblock In \emph{The Eleventh International Conference on Learning Representations}, 2023.
\newblock URL \url{https://openreview.net/forum?id=OKcJhpQiGiX}.

\bibitem[Roth et~al.(2024)Roth, Thede, Koepke, Vinyals, Henaff, and Akata]{roth2024fantastic}
Karsten Roth, Lukas Thede, A.~Sophia Koepke, Oriol Vinyals, Olivier~J Henaff, and Zeynep Akata.
\newblock Fantastic gains and where to find them: On the existence and prospect of general knowledge transfer between any pretrained model.
\newblock In \emph{The Twelfth International Conference on Learning Representations}, 2024.
\newblock URL \url{https://openreview.net/forum?id=m50eKHCttz}.

\bibitem[Sawada and Nakamura(2022)]{sawada2022concept}
Yoshihide Sawada and Keigo Nakamura.
\newblock Concept bottleneck model with additional unsupervised concepts.
\newblock \emph{IEEE Access}, 10:\penalty0 41758--41765, 2022.

\bibitem[Sheth et~al.(2022)Sheth, Rahman, Sevyeri, Havaei, and Kahou]{sheth2022learning}
Ivaxi Sheth, Aamer~Abdul Rahman, Laya~Rafiee Sevyeri, Mohammad Havaei, and Samira~Ebrahimi Kahou.
\newblock Learning from uncertain concepts via test time interventions.
\newblock In \emph{Workshop on Trustworthy and Socially Responsible Machine Learning, NeurIPS 2022}, 2022.

\bibitem[Shin et~al.(2023)Shin, Jo, Ahn, and Lee]{closerlook}
Sungbin Shin, Yohan Jo, Sungsoo Ahn, and Namhoon Lee.
\newblock A closer look at the intervention procedure of concept bottleneck models.
\newblock In Andreas Krause, Emma Brunskill, Kyunghyun Cho, Barbara Engelhardt, Sivan Sabato, and Jonathan Scarlett, editors, \emph{Proceedings of the 40th International Conference on Machine Learning}, volume 202 of \emph{Proceedings of Machine Learning Research}, pages 31504--31520. PMLR, 23--29 Jul 2023.
\newblock URL \url{https://proceedings.mlr.press/v202/shin23a.html}.

\bibitem[Shwartz-Ziv and Tishby(2017)]{shwartzziv2017opening}
Ravid Shwartz-Ziv and Naftali Tishby.
\newblock Opening the black box of deep neural networks via information, 2017.

\bibitem[Wachter et~al.(2018)Wachter, Mittelstadt, and Russell]{legal2}
Sandra Wachter, Brent Mittelstadt, and Chris Russell.
\newblock Counterfactual explanations without opening the black box: Automated decisions and the gdpr, 2018.

\bibitem[Wah et~al.(2011)Wah, Branson, Welinder, Perona, and Belongie]{wah2011caltech}
Catherine Wah, Steve Branson, Peter Welinder, Pietro Perona, and Serge Belongie.
\newblock The caltech-ucsd birds-200-2011 dataset.
\newblock 2011.

\bibitem[Xian et~al.(2018)Xian, Lampert, Schiele, and Akata]{xian2018zero}
Yongqin Xian, Christoph~H Lampert, Bernt Schiele, and Zeynep Akata.
\newblock Zero-shot learning—a comprehensive evaluation of the good, the bad and the ugly.
\newblock \emph{IEEE transactions on pattern analysis and machine intelligence}, 41\penalty0 (9):\penalty0 2251--2265, 2018.

\bibitem[Xu et~al.(2023)Xu, Qin, Mi, Wang, and Li]{xu2023energy}
Xinyue Xu, Yi~Qin, Lu~Mi, Hao Wang, and Xiaomeng Li.
\newblock Energy-based concept bottleneck models.
\newblock In \emph{The Twelfth International Conference on Learning Representations}, 2023.

\bibitem[Yang et~al.(2023)Yang, Panagopoulou, Zhou, Jin, Callison-Burch, and Yatskar]{yang2023language}
Yue Yang, Artemis Panagopoulou, Shenghao Zhou, Daniel Jin, Chris Callison-Burch, and Mark Yatskar.
\newblock Language in a bottle: Language model guided concept bottlenecks for interpretable image classification.
\newblock In \emph{Proceedings of the IEEE/CVF Conference on Computer Vision and Pattern Recognition}, pages 19187--19197, 2023.

\bibitem[Yuksekgonul et~al.(2022)Yuksekgonul, Wang, and Zou]{yuksekgonul2022post}
Mert Yuksekgonul, Maggie Wang, and James Zou.
\newblock Post-hoc concept bottleneck models.
\newblock In \emph{The Eleventh International Conference on Learning Representations}, 2022.

\bibitem[Zarlenga et~al.(2022)Zarlenga, Barbiero, Ciravegna, Marra, Giannini, Diligenti, Shams, Precioso, Melacci, Weller, et~al.]{zarlenga2022concept}
Mateo~Espinosa Zarlenga, Pietro Barbiero, Gabriele Ciravegna, Giuseppe Marra, Francesco Giannini, Michelangelo Diligenti, Zohreh Shams, Frederic Precioso, Stefano Melacci, Adrian Weller, et~al.
\newblock Concept embedding models.
\newblock \emph{arXiv preprint arXiv:2209.09056}, 2022.

\bibitem[Zarlenga et~al.(2023{\natexlab{a}})Zarlenga, Barbiero, Shams, Kazhdan, Bhatt, Weller, and Jamnik]{zarlenga2023towards}
Mateo~Espinosa Zarlenga, Pietro Barbiero, Zohreh Shams, Dmitry Kazhdan, Umang Bhatt, Adrian Weller, and Mateja Jamnik.
\newblock Towards robust metrics for concept representation evaluation.
\newblock \emph{arXiv preprint arXiv:2301.10367}, 2023{\natexlab{a}}.

\bibitem[Zarlenga et~al.(2023{\natexlab{b}})Zarlenga, Collins, Dvijotham, Weller, Shams, and Jamnik]{zarlenga2023learning}
Mateo~Espinosa Zarlenga, Katherine~M Collins, Krishnamurthy~Dj Dvijotham, Adrian Weller, Zohreh Shams, and Mateja Jamnik.
\newblock Learning to receive help: Intervention-aware concept embedding models.
\newblock In \emph{Thirty-seventh Conference on Neural Information Processing Systems}, 2023{\natexlab{b}}.

\end{thebibliography}
}
\newpage

% \section*{Appendix}
\appendix

\section{Details of the Intervention Procedure}
\label{sec:algorithms}

In Algorithm \ref{alg:intervention} we describe the standard process of performing interventions in concept-based models using an intervention policy.

\begin{algorithm}
\caption{Intervention Algorithm}
\label{alg:intervention}
\begin{algorithmic}[1]
\State \textbf{Inputs:}
\State \hspace{\algorithmicindent} $T$ (total number of interventions)
\State \hspace{\algorithmicindent} $\pi$ (intervention policy, which takes the concepts as input)
\State \hspace{\algorithmicindent} $\hat{c}$ (concepts predicted by the concept encoder)

\State $\tilde{c} \gets \hat{c}$ \Comment{output of the concept encoder, g}

\For{$t \in \{0, \ldots, T-1\}$}
    \State $i \gets \pi(\tilde{c})$ \Comment{$i$ is the concept that we want the user to intervene on}
    \State $\tilde{c}_i \gets c_i$ \Comment{replace the $i$th concept in $\tilde{c}$ with its ground truth value $c_i$}
\EndFor

\State \Return $\tilde{y} = f(\tilde{c})$ \Comment{updated class prediction after all interventions have been performed}
\end{algorithmic}
\end{algorithm}

\noindent In Algorithm \ref{alg:realignment_loss} we describe the procedure used in our setup, which realigns unintervened concepts following an intervention step. We use this algorithm to compute the loss for training the realignment model.

\begin{algorithm}
\caption{Realignment Model Training Loss}
\label{alg:realignment_loss}
\begin{algorithmic}[1]
\State \textbf{Inputs:}
\State \hspace{\algorithmicindent} $T$ (total number of interventions)
\State \hspace{\algorithmicindent} $\pi$ (intervention policy, which takes the concepts as input)
\State \hspace{\algorithmicindent} $\hat{c}$ (concepts predicted by the concept encoder)

\State $\tilde{c} \gets \hat{c}$ \Comment{output of the concept encoder, g}
\State $\kappa_{-1} \gets \hat{c}$ \Comment{initialize realigned concepts}
\State $\mathcal{L} \gets 0$ \Comment{initialize loss}

\For{$t \in \{0, \ldots, T-1\}$}
    \State $i \gets \pi(\kappa_{t-1})$ \Comment{$i$ is the concept that we want the user to intervene on}
    \State $\tilde{c}_i \gets c_i$ \Comment{replace the $i$th concept in $\tilde{c}$ with its ground truth value $c_i$}
    \State $\kappa_t \gets u(\tilde{c})$ \Comment{output of realignment model}
    \State $\mathcal{L} \gets \mathcal{L} + \textrm{CE}(\kappa_t, c)$ \Comment{aggregate loss}
\EndFor

\State \Return $\mathcal{L}/T$ \Comment{average loss across all intervention steps}
\end{algorithmic}
\end{algorithm}

\section{Comparison Between Random and UCP Policies}
\label{sec:ucp_vs_random}
\begin{figure}[h]
    \centering
    \begin{subfigure}
        \centering
        \includegraphics[width=0.32\textwidth]{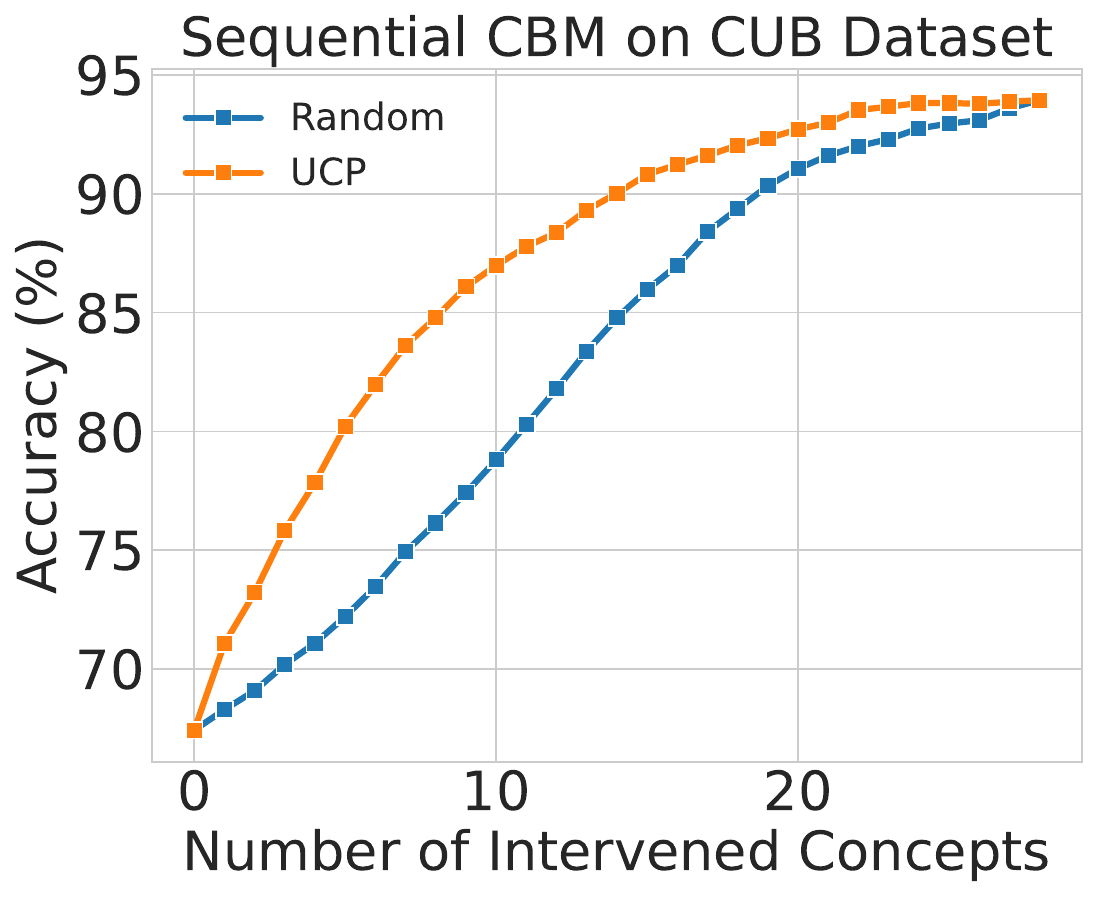}
        % \label{fig:r}
    \end{subfigure}
    \begin{subfigure}
        \centering
        \includegraphics[width=0.32\textwidth]        {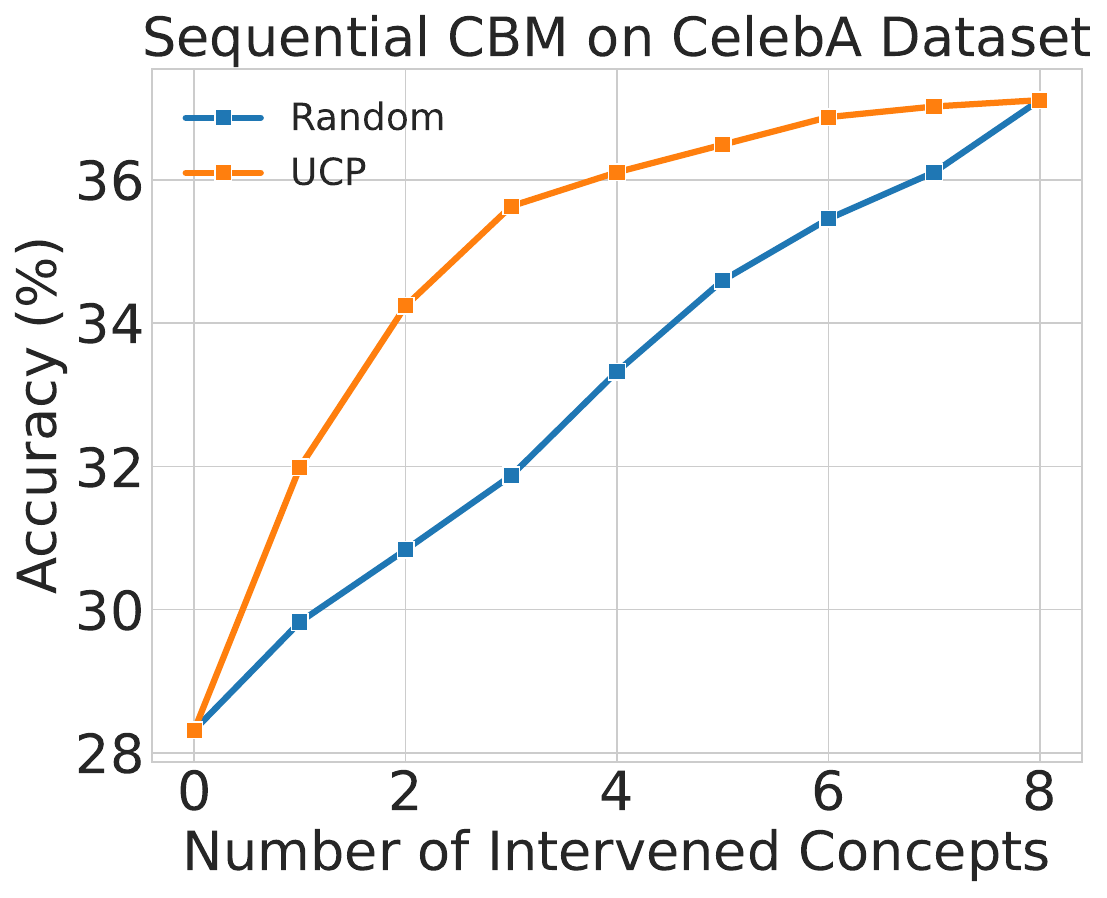}
        % \label{fig:sub2}
    \end{subfigure}
    \begin{subfigure}
        \centering
        \includegraphics[width=0.325\textwidth]        {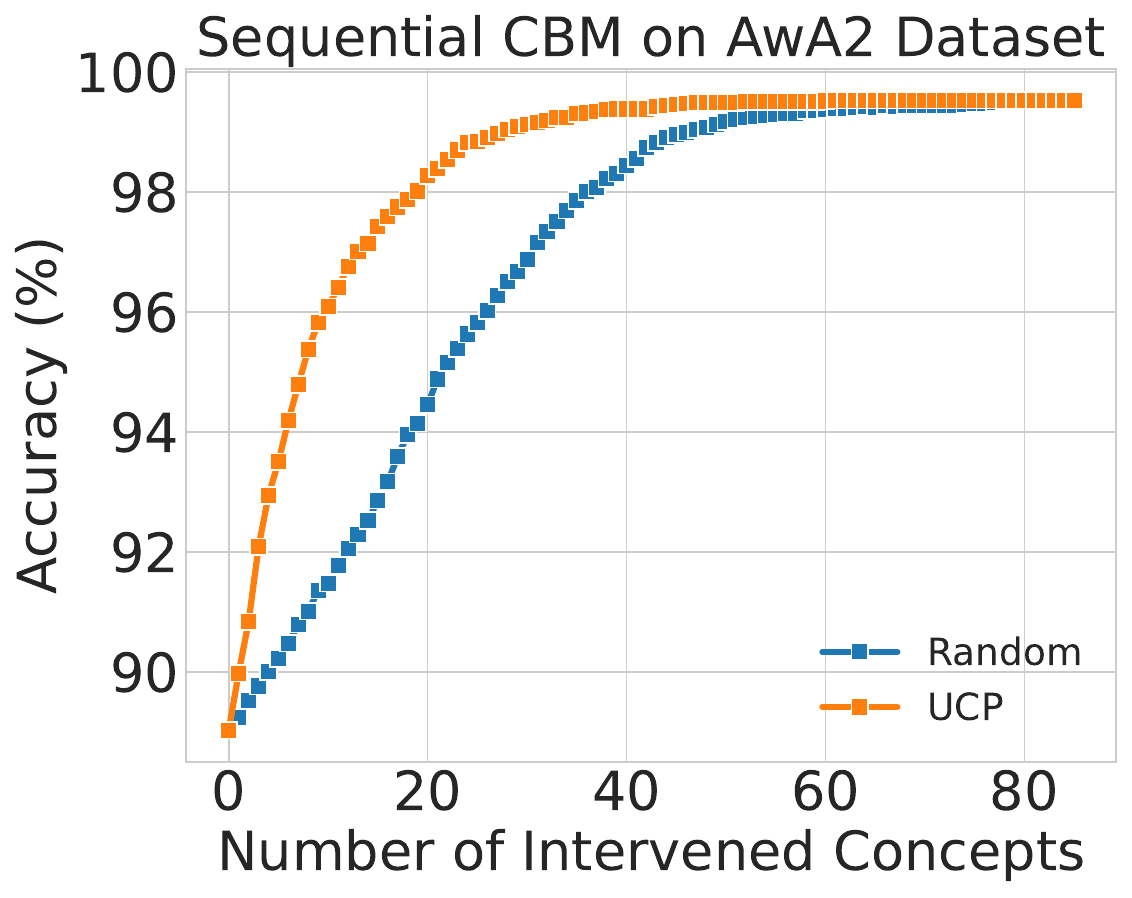}
        % \label{fig:sub3}
    \end{subfigure}
    % \vspace{-15pt}
    \caption{Comparison between accuracy under UCP and Random intervention policies. UCP is superior in all three datasets.}
    \label{fig:ucp_vs_random}
% \vspace{-10pt}    
\end{figure}

In this section, we compare the classification accuracies achieved by following the random and UCP intervention policies on the three datasets, respectively. In Fig. \ref{fig:ucp_vs_random} we show that the UCP policy is superior across all datasets, and is therefore our default policy across all experiments in this study.

\section{Additional Results on IntCEMs}
\label{sec:intcem_appendix}
\begin{figure}[h]
    \centering
    \begin{subfigure}
        \centering
        \includegraphics[width=0.35\textwidth]{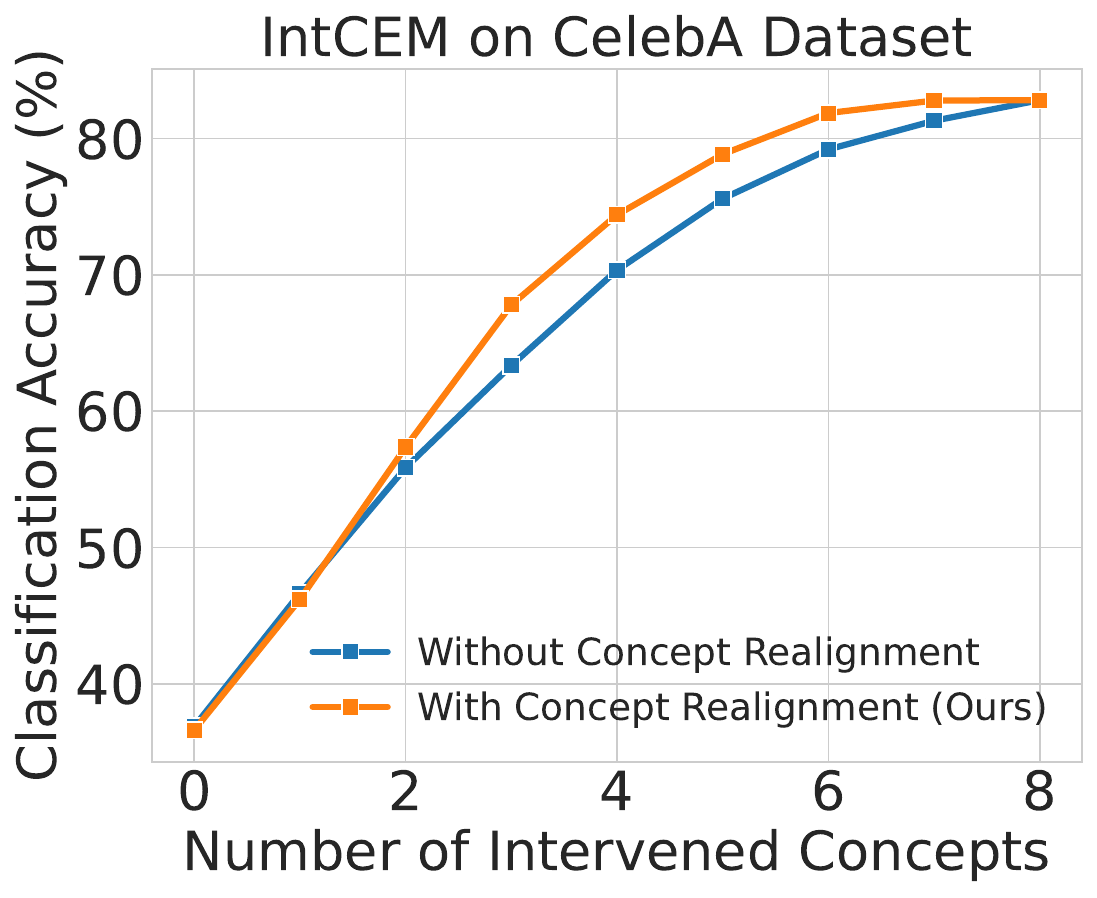}
        % \label{fig:r}
    \end{subfigure}
    \begin{subfigure}
        \centering
        \includegraphics[width=0.36\textwidth]        {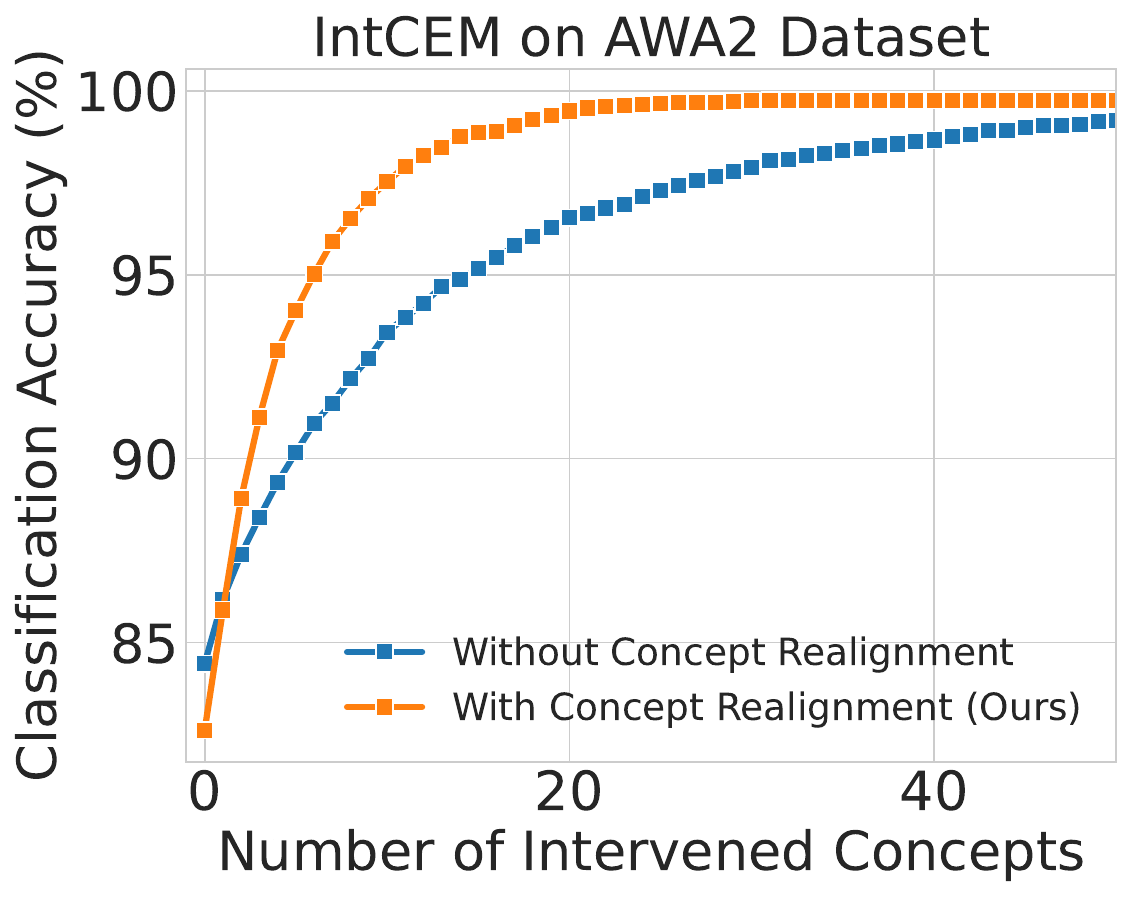}
        % \label{fig:sub2}
    \end{subfigure}
    \caption{Classification accuracy with and without posthoc concept realignment in intervention-aware CEMs. In both cases, concept realignment improves performance of the base IntCEM model.}
    \label{fig:intcem_additional}
\vspace{-10pt}    
\end{figure}

In this section, we report the performance of posthoc concept realignment on the intervention-aware CEMs (IntCEMs) on the CelebA and AwA2 datasets to supplement the results in Section \ref{subsec:joint}. In Fig. \ref{fig:intcem_additional} we show that concept realignment improves the performance of the SoTA approach in both datasets.

\section{Additional Results}
Here, we report the area under the curve of concept prediction loss and classification accuracy using three different random seeds. It can be seen in Tables \ref{tab:3seeds_cub} and \ref{tab:3seeds_celeba} that concept realignment consistently improves performance on both metrics.

\begin{table}[]
\centering
\caption{Area Under Curve (AUC) of Concept Prediction Loss and Classification Accuracy with/without CIRM for three random seeds on the CUB dataset. We use the same backbone for sequential and independent CBMs. CIRM improves performance across all models and runs.}
% \vspace{-5pt}
\resizebox{0.8\linewidth}{!}{%
\begin{tabular}{lccccccc}
\hline
\multirow{2}{*}{\textbf{Base Model}} & \multirow{2}{*}{\textbf{Realigned}} & \multicolumn{3}{c}{\textbf{Concept Loss AUC $\downarrow$}} & \multicolumn{3}{c}{\textbf{Accuracy AUC $\uparrow$}}   \\ \cmidrule{3-8} 
                                     &                                     & Run 1              & Run 2             & Run 3             & Run 1            & Run 2            & Run 3            \\ \hline
\multirow{2}{*}{Sequential CBM}      & $\mathbf{\times}$                   & 6.72               & 7.11              & 6.77              & 2460.86          & 2394.1           & 2444.08          \\
                                     & $\mathbf{\checkmark}$               & \textbf{3.16}      & \textbf{3.16}     & \textbf{3.24}     & \textbf{2510.48} & \textbf{2460.41} & \textbf{2501.08} \\ \hline
\multirow{2}{*}{Independent CBM}     & $\mathbf{\times}$                   & 6.72               & 7.11              & 6.77              & 2653.37          & 2652.75          & 2652.47          \\
                                     & $\mathbf{\checkmark}$               & \textbf{3.16}      & \textbf{3.16}     & \textbf{3.24}     & \textbf{2678.09} & \textbf{2675.04} & \textbf{2675.48} \\ \hline
\multirow{2}{*}{Joint CBM}           & $\mathbf{\times}$                   & 5.93               & 5.84              & 5.89              & 2580.28          & 2533.56          & 2591.32          \\
                                     & $\mathbf{\checkmark}$               & \textbf{3.67}      & \textbf{3.49}     & \textbf{3.58}     & \textbf{2608.89} & \textbf{2559.93} & \textbf{2622.53} \\ \hline
\multirow{2}{*}{CEM}                 & $\mathbf{\times}$                   & 5.99               & 13.19             & 6.50              & 2521.31          & 1681.97          & 2579.84          \\
                                     & $\mathbf{\checkmark}$               & \textbf{3.21}      & \textbf{6.66}     & \textbf{3.43}     & \textbf{2558.07} & \textbf{1762.58} & \textbf{2617.25} \\ \hline
\end{tabular}}
\label{tab:3seeds_cub}
\end{table}

\begin{table}[]
\centering
\caption{Area Under Curve (AUC) of Concept Prediction Loss and Classification Accuracy with/without CIRM for three random seeds on the CelebA dataset. We use the same backbone for sequential and independent CBMs. CIRM improves performance across all models and runs.}
% \vspace{-5pt}
\resizebox{0.8\linewidth}{!}{%
\begin{tabular}{lccccccc}
\hline
\multirow{2}{*}{\textbf{Base Model}} & \multirow{2}{*}{\textbf{Realigned}} & \multicolumn{3}{c}{\textbf{Concept Loss AUC $\downarrow$}} & \multicolumn{3}{c}{\textbf{Accuracy AUC $\uparrow$}} \\ \cline{3-8} 
                                     &                                     & Run 1              & Run 2             & Run 3             & Run 1            & Run 2           & Run 3           \\ \hline
\multirow{2}{*}{Sequential CBM}      & $\mathbf{\times}$                   & 1.59               & 1.64              & 1.65              & 281.09           & 279.64          & 279.47          \\
                                     & $\mathbf{\checkmark}$               & \textbf{1.51}      & \textbf{1.53}     & \textbf{1.55}     & \textbf{284.76}  & \textbf{284.00} & \textbf{284.21} \\ \hline
\multirow{2}{*}{Independent CBM}     & $\mathbf{\times}$                   & 1.59               & 1.64              & 1.65              & 280.86           & 308.38          & 310.57          \\
                                     & $\mathbf{\checkmark}$               & \textbf{1.51}      & \textbf{1.53}     & \textbf{1.55}     & \textbf{282.48}  & \textbf{312.72} & \textbf{316.45} \\ \hline
\multirow{2}{*}{Joint CBM}           & $\mathbf{\times}$                   & 2.88               & 3.23              & 3.10              & 273.06           & 236.22          & 296.80          \\
                                     & $\mathbf{\checkmark}$               & \textbf{1.75}      & \textbf{1.77}     & \textbf{1.74}     & \textbf{273.76}  & \textbf{246.09} & \textbf{303.76} \\ \hline
\multirow{2}{*}{CEM}                 & $\mathbf{\times}$                   & 1.65               & 1.90              & 1.83              & 396.70           & 366.87          & 361.60          \\
                                     & $\mathbf{\checkmark}$               & \textbf{1.49}      & \textbf{1.66}     & \textbf{1.58}     & \textbf{401.84}  & \textbf{370.88} & \textbf{363.57} \\ \hline
\end{tabular}}
\label{tab:3seeds_celeba}
\end{table}

\begin{table}[]
\centering
\caption{Area Under Curve (AUC) of Concept Prediction Loss and Classification Accuracy with/without CIRM for three random seeds on the AwA2 dataset. We use the same backbone for sequential and independent CBMs. CIRM improves performance across all models and runs.}
% \vspace{-5pt}
\resizebox{0.8\linewidth}{!}{%
\begin{tabular}{lcllllll}
\hline
\multirow{2}{*}{\textbf{Base Model}} & \multirow{2}{*}{\textbf{Realigned}} & \multicolumn{3}{c}{\textbf{Concept Loss AUC $\downarrow$}}                        & \multicolumn{3}{c}{\textbf{Accuracy AUC $\uparrow$}}                              \\ \cline{3-8} 
                                     &                                     & \multicolumn{1}{c}{Run 1} & \multicolumn{1}{c}{Run 2} & \multicolumn{1}{c}{Run 3} & \multicolumn{1}{c}{Run 1} & \multicolumn{1}{c}{Run 2} & \multicolumn{1}{c}{Run 3} \\ \hline
\multirow{2}{*}{Sequential CBM}      & $\mathbf{\times}$                   & 4.26                      & 3.76                      & 3.92                      & 8363.9                    & 8411.17                   & 8373.00                   \\
                                     & $\mathbf{\checkmark}$               & \textbf{1.13}             & \textbf{1.2}              & \textbf{1.15}             & \textbf{8397.79}             & \textbf{8437.69}          & \textbf{8400.66}          \\ \hline
\multirow{2}{*}{Independent CBM}     & $\mathbf{\times}$                   & 4.26                      & 3.76                      & 3.92                      & 8403.45                   & 8410.39                   & 8407.77                   \\
                                     & $\mathbf{\checkmark}$               & \textbf{1.13}             & \textbf{1.2}              & \textbf{1.15}             & \textbf{8437.31}          & \textbf{8437.59}          & \textbf{8438.71}          \\ \hline
\multirow{2}{*}{Joint CBM}           & $\mathbf{\times}$                   & 4.77                      & 4.34                      & 4.47                      & 8276.37                   & 8350.96                   & 8346.31                   \\
                                     & $\mathbf{\checkmark}$               & \textbf{1.5}              & \textbf{1.54}             & \textbf{1.51}             & \textbf{8326.95}          & \textbf{8391.89}          & \textbf{8389.76}          \\ \hline
\multirow{2}{*}{CEM}                 & $\mathbf{\times}$                   & 4.9                       & 4.04                      & 3.92                      & 8429.35                   & 8438.99                   & 8439.87                   \\
                                     & $\mathbf{\checkmark}$               & \textbf{1.69}             & \textbf{1.45}             & \textbf{1.46}             & \textbf{8433.38}          & \textbf{8439.9}           & \textbf{8439.52}          \\ \hline
\end{tabular}}
\label{tab:3seeds_awa2}
\end{table}

\end{document}